\pdfoutput=1
\documentclass[10pt,twocolumn,letterpaper]{article}

\usepackage[pagenumbers]{cvpr} %

\usepackage{xcolor}
\usepackage{caption} %

\usepackage{xspace}
\usepackage{booktabs}
\usepackage{colortbl} %
\usepackage{multirow}
\usepackage{amsmath,amssymb}
\usepackage{pifont}
\usepackage{listings} %
\newcommand{\ours}{LesionAgent\xspace}
\newcommand{\cmark}{\ding{51}}
\newcommand{\tool}[1]{\texttt{\small #1}}

\definecolor{cvprblue}{rgb}{0.21,0.49,0.74}
\usepackage[pagebackref,breaklinks,colorlinks,allcolors=cvprblue]{hyperref}

\def\confName{CVPR}
\def\confYear{2027}
\title{\ours: Training-Free Agentic Tumor Tracking and Response Assessment\\in Longitudinal CT}

\author{Maximilian Rokuss\textsuperscript{1,2},
Jeya Maria Jose Valanarasu\textsuperscript{4},
Alejandro Lozano\textsuperscript{5},
Benjamin Hamm\textsuperscript{1,3},\\
Yannick Kirchhoff\textsuperscript{1,2,6},
Fabian Isensee\textsuperscript{1,7},
Klaus Maier-Hein\textsuperscript{1,2,3,7,8}
\\
\and
\parbox{\dimexpr\textwidth-2\tabcolsep}{\centering\small
\textsuperscript{1}Division of Medical Image Computing, German Cancer Research Center (DKFZ), Heidelberg, Germany\\
\textsuperscript{2}Faculty of Mathematics and Computer Science and \textsuperscript{3}Medical Faculty, Heidelberg University, Germany \quad
\textsuperscript{4}Microsoft Research\\
\textsuperscript{5}Stanford University \quad
\textsuperscript{6}HIDSS4Health, Heidelberg \quad
\textsuperscript{7}National Center for Tumor Diseases (NCT) Heidelberg, Germany\\
\textsuperscript{8}Pattern Analysis and Learning Group, Department of Radiation Oncology, Heidelberg University Hospital, Germany}
\\[2pt]
{\tt\small maximilian.rokuss@dkfz-heidelberg.de}
}

\begin{document}
\twocolumn[{\renewcommand\twocolumn[1][]{#1}\maketitle\begin{center}
\centering
\includegraphics[width=\textwidth]{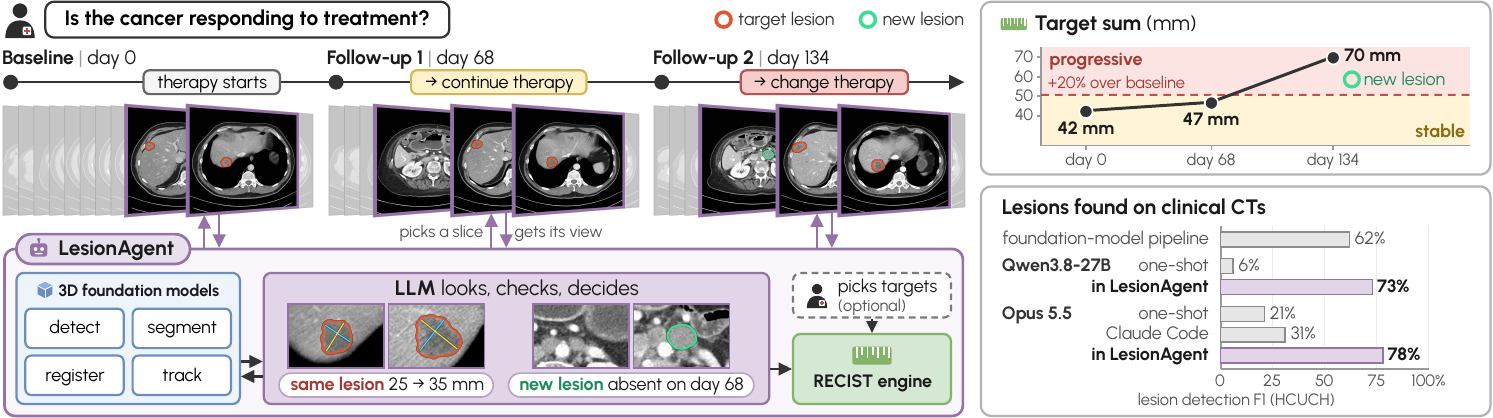}
\captionof{figure}{\textbf{\ours assesses a patient's treatment response over time, without any task-specific training.} This requires interpreting several CT scans and applying the many rules of a clinical guideline, which no single model or fixed pipeline does reliably. \ours instead operates frozen 3D foundation models as tools, checks their outputs on selected slices (purple frames) and follows every lesion across visits. A deterministic RECIST engine turns its measurements into the response: growing targets and a new lesion indicate progression (top right). Inside the \ours harness, off-the-shelf LLMs find far more lesions than on their own (bottom right).}
\label{fig:teaser}
\end{center}\vspace{1mm}
}]
\begin{abstract}
Every few months, a cancer patient's CT scans decide whether treatment continues. 3D foundation models can now detect, segment, register and track lesions, yet none of them reads the scans as a radiologist does. A radiologist decides which findings are lesions, which are the same lesion months later and whether a new spot means progression, and checks each decision against the extensive rules of a clinical guideline. We introduce \ours, an LLM agent that performs this read end to end without task-specific training. It operates frozen 3D models as tools, acts in 3D through coordinate-labelled views, remembers every lesion across visits in a registry, verifies every model output it relies on, and leaves every number to a deterministic guideline engine. On four public longitudinal CT datasets, \ours tracks more lesions and measures them more accurately than every automatic baseline, and where the same 3D models chained by fixed rules call progression falsely at 33 of 56 follow-ups, it does so twice. On clinical follow-up CTs, it agrees with the radiologists' response at 20 of 23 follow-ups, including every change in response, and without training on DeepLesion, it retrieves more of its lesions than every tracker trained on it (89.0\% vs.\ 87.4\%). Its harness, not its LLM, makes the difference: given the same LLM and 3D models but not the harness, Claude Code reaches a detection F1 of 41\% and \ours 78\%, and an open-weight LLM that can run inside the hospital rises from 6\% to 73\%. LesionAgent and a napari reading app for any LLM are available at \url{https://github.com/MIC-DKFZ/LesionAgent}.
\end{abstract}

\section{Introduction}
\label{sec:intro}

Patients with cancer are scanned every few months, and at every scan a radiologist decides whether the disease
responds, is stable or progresses. This decision determines whether the treatment is continued, modified or
changed. Under the clinical guideline RECIST~1.1~\cite{eisenhauer2009recist}, this \emph{read} means a)~finding all lesions in a
3D volume, b)~selecting up to five target lesions and measuring them,
c)~matching each lesion across scans with different fields of view, contrast and anatomy, d)~deciding whether a new finding is a new lesion, and e)~applying the
guideline's rules. The read is manual and slow, and it varies between readers: two radiologists choose the same target lesions in only 41\% of patients~\cite{kuhl2019recist} and often disagree on the response~\cite{karmakar2019}. Small lesions are easily missed, even by
radiologists, whose sensitivity for lung nodules ranges from 51 to 83\%~\cite{armato2009nodules}. 

\noindent 3D foundation models now solve several of these perception steps zero-shot, one at a time. Text-promptable models detect and segment lesions named in natural language~\cite{voxtell}, promptable models segment them from a click~\cite{nninteractive}, and registration and longitudinal models find a tumor again in a follow-up scan~\cite{unigradicon,longitrack,lesionlocator}. None of them, however, performs the whole response assessment. It requires composing these models, catching their failures and making decisions no model is trained for: is a finding a lesion, is it the same lesion months later, is a new finding real, and which lesions should be followed? Learning all of this end to end would need longitudinal annotations at a scale that does not exist, as even recent releases cover a single cancer type~\cite{kuestner2025longitudinalct,longitrack} or a few dozen patients~\cite{hcuch}.\\

\noindent LLM agents compose pretrained models without training: they call models as tools
and solve compositional visual tasks~\cite{visprog,vipergpt,react}, with medical agents calling specialist models for single images or studies~\cite{medrax,ctagent,radagent,3dmedagent}. 
However, none of these agents assesses response across a patient's CT scans. Current multimodal LLMs cannot take several 3D volumes~\cite{3drad,suprijadi2026resolution}, and even frontier agents that operate a clinical viewer find almost no new lesions in follow-up CT, failing at perception rather than tool use~\cite{abra,mmbu}. An agent for this task must act in 3D while it sees only 2D images, remember every lesion across visits over a read of many tool calls, verify model outputs that it can only inspect slice by slice, since a single wrong mask can flip the response, and apply the guideline's rules exactly.\\

\noindent We introduce \ours, an LLM agent that performs this read end to end, from the CT scans to the response at every visit, without any task-specific training. At baseline, a text-promptable model proposes lesion candidates, and the agent accepts or rejects each one in rendered views. Every accepted lesion enters a registry (lesions $\times$ visits) that holds its mask at every visit. The agent chooses the target lesions itself or, if a
clinician is available, lets them pick from its best valid sets. At each follow-up, registration carries every lesion into the new scan, where a longitudinal model segments it. If nothing is found, the agent inspects the registered position and records the lesion as absent. A finding that matches no tracked lesion becomes a new lesion only after the agent has checked the same position at baseline. Renders are labelled with voxel coordinates, so the agent turns a position it sees into a 3D tool call, \eg a click on a lesion that detection missed, and a second model re-segments every mask to catch failures. A deterministic guideline engine measures the lesions on their masks and applies RECIST, so the LLM never computes a number or a category. Another guideline would simply change the engine, not the agent.\\

\noindent We evaluate \ours end to end on four public longitudinal CT datasets, fully automatically or with one click per lesion, with identical settings on every dataset. Since a response is only as reliable as the lesions behind it, we score the detection, tracking and measurement of every lesion as well as the response, against chains of the same 3D models, trained trackers, Claude Code and one-shot LLMs, and against radiologists. Fully automatically, \ours tracks more lesions than every baseline on all four datasets, and where the same 3D models chained by fixed rules call progression falsely at 33 of 56 follow-ups, it does so twice. On clinical follow-up CTs, it agrees with the radiologists' response at 20 of 23 follow-ups, including all four changes in response (95\% balanced accuracy). Its harness, not its LLM, makes the difference: given the same LLM and 3D models on these CTs, Claude Code reaches a detection F1 of 41\% and \ours 78\%, and an open-weight LLM that can run inside the hospital rises from 6\% to 73\%. In summary, our contributions are:

\begin{itemize}
\item \textbf{Training-Free Agentic Tumor Monitoring:} \ours is the first agent that reads longitudinal CT end to end, from the images to the RECIST response, using only frozen 3D foundation models. It runs fully automatically or with clinician interaction inside a napari viewer plugin.
\item \textbf{A Harness for 3D\,+\,Time:} We let an LLM act in 3D through coordinate-labelled views, remember every lesion across visits in a registry, verify the model outputs it relies on, and leave every number and category to a deterministic guideline engine.
\item \textbf{Clinically Grounded Reads without Training:} \ours agrees with the radiologists' response on clinical follow-up CTs, measures nodules within their inter-reader variability and retrieves lesions across time better than the trackers trained on DeepLesion. Every response it gives rests on lesions a clinician can check.
\end{itemize}

\section{Related Work}
\label{sec:related}

\paragraph{3D foundation models for medical images.}Promptable segmentation has moved from natural images (SAM~\cite{sam}) to medical adaptations like MedSAM~\cite{medsam}, SAM-Med3D~\cite{sammed3d}, SegVol~\cite{segvol} and nnInteractive~\cite{nninteractive}, with text-prompted models such as SAT~\cite{sat}, BiomedParse~\cite{biomedparse} and VoxTell~\cite{voxtell} segmenting structures named in free text. TotalSegmentator~\cite{totalsegmentator} remains the standard for organs, and dedicated models segment lesions~\cite{uls23,lesionlocator,nnfoundation}. Registration models such as ConvexAdam~\cite{convexadam} and uniGradICON~\cite{unigradicon} are used widely to align a patient's scans across time. However, each of these models solves one step of tumor monitoring.

\paragraph{Longitudinal lesion tracking.}
Trackers either predict where a known lesion lies at follow-up or segment it there. Point-based trackers return its position: DLT~\cite{dlt} and TLT~\cite{tlt} retrieve lesion centers on the DeepLesion benchmark~\cite{deeplesion}, and compare against VoxelMorph~\cite{voxelmorph} and DEEDS~\cite{deeds}
registration. Volumetric trackers return the lesion's follow-up mask, segmented after registration~\cite{hering2021wholebody} or conditioned on the earlier scan~\cite{longiseg,kirchhoff2025promptable}.
LongiTrack~\cite{longitrack} combines a registration-propagated point with the lesion's baseline appearance to stay robust where single-time-point models degrade because registration displaces the point prompt~\cite{rocholl2025unstable}. All of them follow a lesion that a radiologist has chosen and leave it to the reader which lesions to follow, whether a finding is new and what a change means.

\paragraph{Radiology agents.} Agents that act through tools~\cite{react} or write programs over pretrained vision modules~\cite{visprog,vipergpt} answer compositional visual queries without training. In radiology, tool-calling agents select among specialist models across modalities~\cite{mmedagent}, for chest X-ray~\cite{medrax} or for chest CT~\cite{ctagent,radagent}. In 3D, VoxelPrompt~\cite{voxelprompt} generates code that calls a jointly trained network, multimodal agents call 3D segmentation models~\cite{3dmedagent,nninteractiveagent}, and DeepTumorVQA~\cite{deeptumorvqa} benchmarks tool-augmented agents on CT questions. All of them answer one query about a single study. Current multimodal LLMs cannot take several 3D volumes, so even a longitudinal CT benchmark gives the prior scans as labels~\cite{3drad}. Frontier agents that operate a clinical viewer were shown to find almost no new lesions in follow-up CT~\cite{abra}, with imaging as their hardest domain overall~\cite{healthagentbench,medopenclaw}. So far, agents for
follow-up imaging assess only brain tumors on MRI: the language model assigns the response category itself~\cite{erdur2026agentic}, or fixed rules score tumor volumes from a segmentation network~\cite{jabal2026btrads}. To our knowledge, no agent has yet assessed tumor response across longitudinal CT, where many lesions in different organs must be found, matched across scans taken months apart and measured. \ours does so without task-specific training, and every response it gives rests on lesions it has found, tracked and measured itself.

\paragraph{Response assessment and reader variability.}
Response criteria turn a patient's scans into a decision on treatment. RECIST~1.1~\cite{eisenhauer2009recist}, the
standard for solid tumors in clinical trials, sums the diameters of up to five target lesions and maps their change,
together with new lesions, to a response category. Image-based RECIST automation addresses single steps such as diameter
measurement~\cite{tang2018recist} or the
tracking and measurement of target lesions a reader has selected~\cite{dahm2024automated}, and language models classify
response from the text of the radiologist's report~\cite{mergen2026llm,leszczynska2026review}. In both cases, a radiologist still performs the read. Even between expert readers, this read varies widely: the inter-reader 95\% limits of agreement for the change of a single lesion are near $\pm30\%$~\cite{yoon2016variability}, and readers choose the same target lesions in only 41\% of patients~\cite{kuhl2019recist}. We therefore compare the agent with human readers not only on the final response but also on each step of the read.

\begin{figure*}[t]
\centering
\includegraphics[width=\linewidth]{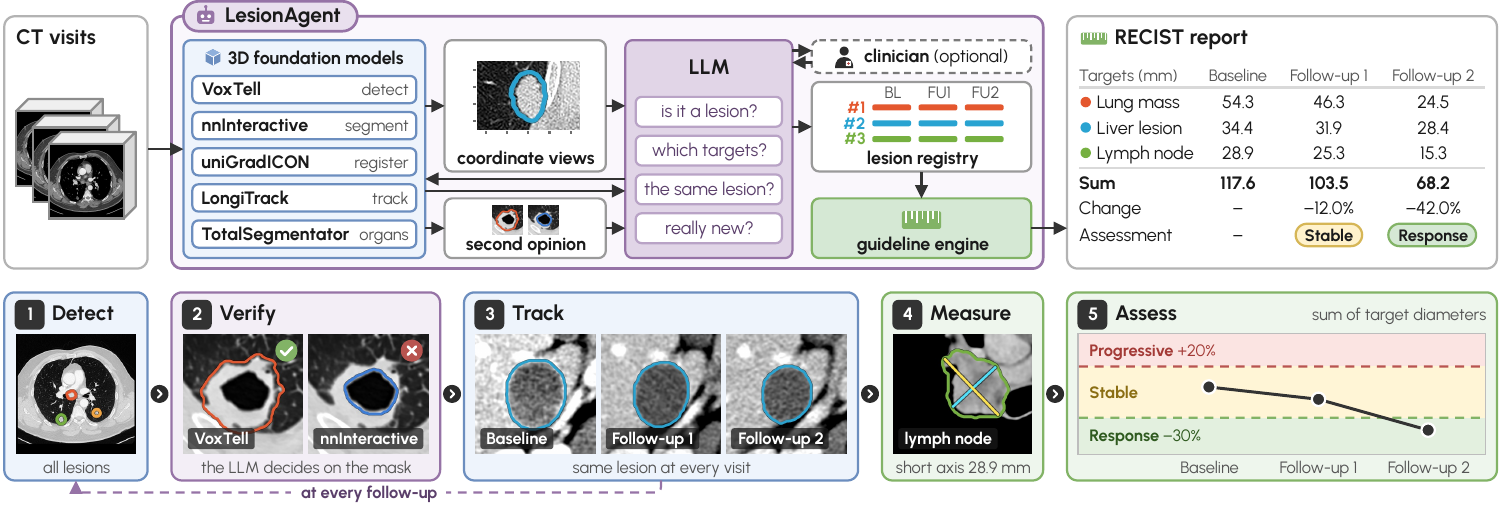}
\caption{\textbf{\ours.} An LLM calls frozen 3D foundation models as tools (top). It checks their outputs in coordinate-labelled views, keeps every lesion at every visit in a registry (BL = baseline, FU = follow-up) and verifies masks with a second opinion. A deterministic guideline engine measures the lesions, assigns the response and writes the RECIST report. Optionally, a clinician answers the LLM's questions (dashed). The bottom row shows the five steps of a read on real CT crops, from detecting all lesions to assessing the response.}
\label{fig:framework}
\end{figure*}

\section{LesionAgent}
\label{sec:method}

Given a patient's CT scans from several visits, \ours finds, tracks and measures the lesions and assesses the
response at every follow-up (\cref{fig:framework}). We first define this task, the \emph{read}, and then describe the
agent, its tools and lesion registry, the views through which it acts in 3D, its self-verification and the selection
of target lesions.

\paragraph{The read.} A patient study is a series of CT scans $s_1,\dots,s_T$, one per visit, with baseline $s_1$. At every follow-up,
RECIST~1.1~\cite{eisenhauer2009recist} compares the sum of the diameters of up to five target lesions with its
baseline value and with its smallest value so far. A decrease of 30\% is a \emph{partial response} and the
disappearance of all lesions a \emph{complete response}. An increase of 20\% and 5\,mm, a new lesion or clear growth of
the other lesions is \emph{progression}, and everything else is \emph{stable disease}. The first progression dates the time to progression, a common endpoint of oncology trials.

\noindent \ours splits the read by the kind of decision. 3D foundation models perceive the lesions. The LLM judges what the tools cannot, namely whether a finding is a lesion, which lesions are targets and whether a new finding is
certainly new, which RECIST calls unequivocal. A deterministic guideline engine computes every number and category.

\paragraph{Agent.}
\label{sec:overview}
\ours is an LLM in a tool-use loop (\cref{fig:framework}). At every step it sees the task and all earlier tool calls
with their results, text and rendered images, and chooses the next call. It acts only through 18 typed tools
(\cref{tab:tools}) on a shared workspace with the scans, the lesion registry, the 3D models and the guideline engine.
Without shell, file access or code execution, it cannot change anything outside its read, and with an open-weight LLM
on local hardware no image leaves the hospital. A short \emph{skill} document (\cref{sec:supp-skill}) describes the
reading workflow, the RECIST rules and when to check a result. Claude Opus~5.5 and the open-weight Qwen3.8-27B use it unchanged.

\begin{table}[t]
\centering
\caption{\textbf{The 18 tools}, the agent's only means of action.}
\label{tab:tools}
\small
\setlength{\tabcolsep}{3pt}
\begin{tabular}{@{}lp{0.72\linewidth}@{}}
\toprule
Group & Tools (model) \\
\midrule
Perception & \tool{detect} (VoxTell), \tool{segment} (nnInteractive / LongiTrack), \tool{track\_all} (uniGradICON +
  LongiTrack), \tool{refine\_lesion} (nnInteractive), \tool{edit\_mask}, \tool{totalsegmentator} (TotalSegmentator) \\
Inspection & \tool{study\_info}, \tool{view}, \tool{view\_pair}, \tool{inspect} \\
Registry & \tool{review\_candidates}, \tool{update\_lesion}, \tool{set\_visit\_judgement} \\
Decision & \tool{score\_lesions}, \tool{propose\_targets}, \tool{set\_targets}, \tool{recist\_report} (engine) \\
Human & \tool{ask\_clinician} \\
\bottomrule
\end{tabular}
\end{table}

\paragraph{Tools.} \tool{detect} proposes lesion candidates using VoxTell~\cite{voxtell} with multiple prompts, and \tool{segment}
adds a lesion that detection missed, or redoes a wrong tracked mask, from one nnInteractive click~\cite{nninteractive}. At a follow-up,
\tool{track\_all} registers the scans with uniGradICON~\cite{unigradicon}, propagates each lesion's point and segments
the lesion there with LongiTrack~\cite{longitrack}, its previous mask serving as context. \tool{detect} then returns only findings that no tracked lesion covers, as candidates for new lesions (all tools in \cref{sec:supp-tools}).

\paragraph{Lesion registry.} The state of the read lives outside the LLM, in a registry of lesions $\times$ visits. Each lesion has a type, an organ, a role (target, non-target, new or rejected) and, at every visit, a mask or a status such as too small to measure, absent or merged with another lesion. Every tool reads and writes the registry, and the report lists every
baseline lesion that still lacks a status at a follow-up, so no lesion is silently lost.

\paragraph{Acting in 3D through 2D views.}
\label{sec:interface}
The LLM sees 2D images, while the tools act in 3D. We therefore label the axes of every view with absolute voxel
indices and patient directions, extending coordinate prompting~\cite{lei2024scaffold,yang2023setofmark} from 2D
images to 3D volumes~\cite{nninteractiveagent}. A position the agent reads off a view can thus become a 3D prompt, \eg a click on a lesion that
detection missed. Every lesion in the registry is outlined in its own colour and labelled with its id at every visit,
optionally with its long and short axis, so the agent can tell which lesion it is looking at, compare it across visits
and check its measurement. Instead of fixed slices, the agent requests the view that the current decision needs, such
as a)~slices or a montage through one lesion to check its mask in 3D, b)~one lesion side by side at several visits
with its registered position to check the tracking, c)~a new finding next to the same position at baseline to tell it
from a missed baseline lesion, or d)~maximum-intensity projections of consecutive 10\,mm slabs to find lesions that
detection missed, since lung nodules appear there as dots and vessels as lines.

\begin{figure*}[t]
\centering
\includegraphics[width=0.985\linewidth]{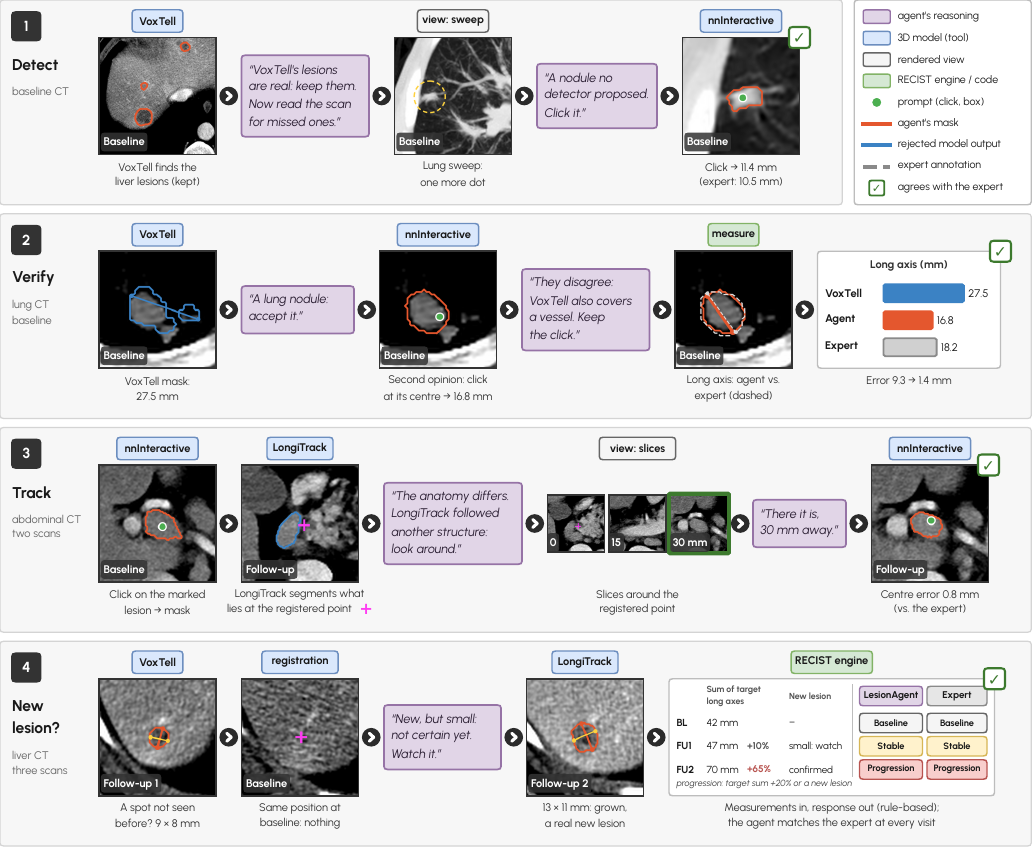}
\caption{\textbf{What \ours catches.} One real case per step of a read, from our test data: the tools' outputs, the
agent's reasoning (shortened from the run log) and the prompts it gives nnInteractive. (1)~The agent keeps the lesions
VoxTell finds and, reading the scan itself, adds a nodule that no detector proposed. (2)~A second opinion exposes a mask that includes a vessel.
(3)~LongiTrack follows another structure at the registered point, and the agent looks at the slices around it. (4)~A small new spot counts as a new lesion only once it has grown, and the guideline engine derives the response. Every result agrees with the expert.}
\label{fig:traces}
\end{figure*}

\paragraph{Self-verification.}
\label{sec:verification}
The 3D models can fail silently. VoxTell masks can be far too small, LongiTrack may follow a neighbouring structure
and registration may land next to a lesion (\cref{fig:traces}). The tools therefore check every output in code and warn the agent when a check fails. Every VoxTell mask the agent accepts (\tool{review\_candidates}) and every tracked mask gets a second opinion from one nnInteractive click at its centre. If the long axes differ by more than
$\max(5\,\mathrm{mm}, 30\%)$, about the inter-reader range~\cite{yoon2016variability}, the agent sees both masks side
by side and chooses. A tracked mask is flagged for review across visits (\tool{view\_pair}) if it is empty,
changes its volume more than threefold, misses the propagated point, splits or touches another lesion. Since most
false new lesions are baseline lesions that detection missed, \tool{set\_visit\_judgement} shows every new lesion next
to the same position at baseline and lists reasons for doubt, such as a size below 10\,mm, no growth or a nearby
baseline finding. If any applies, the agent must state why it is certain. Finally, the agent re-checks visits
within 5\,mm of a response threshold, marked by \tool{recist\_report}, and any progression due to a single new lesion.

\paragraph{Target selection and the guideline engine.}
\label{sec:selection}
RECIST asks for up to five target lesions, at most two per organ, that are large, represent all involved organs and can be measured reproducibly. Code can check size and these constraints, but reproducibility has to be judged from the images. The agent therefore scores how reliably each measurable baseline lesion can be measured again ($q_i\in[0,1]$), lowering the score for ill-defined margins, necrosis or confluence (\tool{score\_lesions}). Code then enumerates all valid target sets $S$ and ranks them by the number of organs covered, then by $\sum_{i\in S} q_i\,s_i$, with $s_i$ the diameter of lesion $i$ relative to the largest (\tool{propose\_targets}). The agent takes the best set or justifies another, or lets a clinician choose, \eg in our napari reading application (\cref{sec:supp-app}). The guideline
engine measures every lesion on its mask and applies RECIST~1.1 with all its conventions, \eg short axes for lymph
nodes and 5\,mm for a target too small to measure. Another guideline, such as iRECIST~\cite{irecist},
mRECIST~\cite{mrecist}, RANO~\cite{rano} or the Lugano criteria~\cite{lugano}, would change the engine, not the agent.

\section{Experiments}
\label{sec:experiments}

\begin{table*}[t]
\centering
\caption{\textbf{End-to-end lesion tracking and response assessment.} \emph{Interactive} methods get one click per
baseline lesion, \emph{automatic} methods only the images. Without task-specific training, \ours is best or on par on
every metric in both settings, and with only open-weight models it is still the best open method on most metrics.
Metrics are in \% except LD (mm), Resp.\ is a balanced accuracy and $\pm$
the bootstrap standard deviation. \textbf{Bold} marks the best per setting, \underline{underlined} the best open-weight automatic method.
}
\label{tab:main}
\small
\providecommand{\sd}[1]{{\tiny$\pm$#1}}
\setlength{\tabcolsep}{2.1pt}
\begin{tabular}{@{}cllcccccccccccccc@{}}
\toprule
 & & & \multicolumn{4}{c}{HCUCH~\cite{hcuch}} & \multicolumn{4}{c}{TumorMap~\cite{tumormap}} & \multicolumn{3}{c}{PanTrack~\cite{longitrack}} & \multicolumn{3}{c}{NCI-MATCH~\cite{eay131}} \\
\cmidrule(lr){4-7}\cmidrule(lr){8-11}\cmidrule(lr){12-14}\cmidrule(l){15-17}
 & & Method & F1 & Track & LD\,$\downarrow$ & Resp. & F1 & Track & LD\,$\downarrow$ & Resp. & Rec. & Track & LD\,$\downarrow$ & Rec. & Track & LD\,$\downarrow$ \\
\midrule
 & \multirow{3}{*}{\textit{Models}} & LesionLocator~\cite{lesionlocator} & -- & 98\sd{2} & 2.8\sd{0.4} & 75\sd{15} & -- & 74\sd{4} & 7.3\sd{1.3} & 29\sd{9} & -- & 94\sd{3} & 3.3\sd{1.0} & -- & 81\sd{7} & 4.0\sd{1.2} \\
 & & LongiTrack~\cite{longitrack} & -- & 98\sd{2} & 3.0\sd{0.6} & 91\sd{5} & -- & 73\sd{4} & 7.9\sd{1.6} & 29\sd{6} & -- & 92\sd{3} & 3.0\sd{0.4} & -- & 74\sd{8} & 3.9\sd{1.0} \\
 & & Reg.$\to$ULS~\cite{uls23} & -- & 98\sd{2} & 3.9\sd{0.5} & \textbf{95}\sd{4} & -- & 85\sd{3} & 7.9\sd{2.4} & 62\sd{10} & -- & 83\sd{5} & 3.8\sd{0.9} & -- & 84\sd{6} & 5.5\sd{1.9} \\
\cmidrule(lr){2-17}
\rowcolor{gray!15}\cellcolor{white}\multirow{-4}{*}{\rotatebox{90}{\scriptsize\textsc{Interactive}}}
 & \cellcolor{white}\textit{Agentic} & \textbf{\ours} (Opus~5.5) & -- & \textbf{100}\sd{0} & \textbf{1.6}\sd{0.4} & \textbf{95}\sd{4} & -- & \textbf{91}\sd{3} & \textbf{4.5}\sd{0.8} & \textbf{69}\sd{9} & -- & \textbf{97}\sd{2} & \textbf{2.6}\sd{0.3} & -- & \textbf{97}\sd{3} & \textbf{3.3}\sd{1.2} \\
\midrule
 & \textit{Pipeline} & VoxTell$\to$LongiTrack & 62\sd{2} & \underline{86}\sd{5} & 4.5\sd{0.8} & 40\sd{13} & \underline{52}\sd{3} & 54\sd{5} & \underline{8.1}\sd{1.9} & 37\sd{7} & \textbf{\underline{93}}\sd{6} & 84\sd{6} & 12.7\sd{3.4} & 52\sd{9} & 39\sd{8} & \underline{4.8}\sd{3.1} \\
\cmidrule(lr){2-17}
 & \multirow{3}{*}{\textit{\shortstack{LLM\\one-shot}}} & Qwen3.8~\cite{qwen38} & 6\sd{2} & 14\sd{5} & 12.6\sd{3.8} & 28\sd{8} & 18\sd{4} & 10\sd{3} & 13.0\sd{3.2} & 40\sd{9} & 0\sd{0} & 0\sd{0} & -- & 13\sd{6} & 10\sd{5} & 16.9\sd{5.5} \\
 & & MedGemma 1.5~\cite{medgemma15} & 3\sd{2} & 0\sd{0} & -- & 18\sd{6} & 0\sd{0} & 0\sd{0} & -- & 35\sd{9} & 0\sd{0} & 0\sd{0} & -- & 0\sd{0} & 0\sd{0} & -- \\
 & & Opus~5.5~\cite{opus55} & 21\sd{3} & 53\sd{7} & 6.2\sd{1.1} & 88\sd{6} & 40\sd{4} & 27\sd{4} & 8.4\sd{2.0} & 69\sd{10} & 53\sd{13} & 22\sd{7} & 12.0\sd{2.5} & 42\sd{9} & 39\sd{9} & 9.1\sd{2.8} \\
\cmidrule(lr){2-17}
 & & Claude Code & 31\sd{4} & 63\sd{7} & 2.4\sd{0.4} & 80\sd{14} & 50\sd{4} & 37\sd{4} & 4.4\sd{0.9} & 74\sd{9} & 60\sd{13} & 54\sd{8} & 11.1\sd{1.8} & 36\sd{9} & 29\sd{8} & 6.7\sd{6.0} \\
\rowcolor{gray!15}\cellcolor{white} & \cellcolor{white} & \textbf{\ours} (Opus~5.5) & \textbf{78}\sd{2} & \textbf{88}\sd{5} & \textbf{1.9}\sd{0.6} & \textbf{95}\sd{3} & \textbf{62}\sd{3} & \textbf{75}\sd{4} & \textbf{4.0}\sd{0.5} & \textbf{79}\sd{6} & \textbf{93}\sd{6} & \textbf{97}\sd{3} & \textbf{5.9}\sd{1.4} & 55\sd{9} & 55\sd{9} & \textbf{2.8}\sd{1.7} \\
\rowcolor{gray!7}\cellcolor{white}\multirow{-7}{*}{\rotatebox{90}{\scriptsize\textsc{Automatic}}}
 & \cellcolor{white}\multirow{-3}{*}{\textit{Agentic}} & \ours\ (Qwen3.8) & \underline{73}\sd{2} & 82\sd{6} & \underline{4.4}\sd{1.5} & \underline{62}\sd{14} & 50\sd{3} & \underline{58}\sd{4} & 10.3\sd{2.4} & \underline{52}\sd{11} & 87\sd{9} & \underline{89}\sd{5} & \underline{9.6}\sd{2.5} & \textbf{\underline{61}}\sd{9} & \textbf{\underline{58}}\sd{9} & 6.9\sd{3.0} \\
\bottomrule
\end{tabular}
\end{table*}

We evaluate \ours along four axes: (i)~the full task end to end, from the images to the response at every visit,
against tool chains, specialised trackers, Claude Code and one-shot LLMs; (ii)~lesion tracking in isolation against trained trackers; (iii)~each step against human readers; and (iv)~the contribution of
each capability. The agent's tools are used as released: none of the four longitudinal datasets is in their or the baselines'
training data, nor was any tool trained on DeepLesion tracking. Skill development and all ablations were run on HCUCH.

\paragraph{Data.} All datasets are public (details in \cref{sec:supp-data}).
\textbf{HCUCH}~\cite{hcuch}: clinical RECIST~1.1 target measurements on thorax and abdomen CT of 14 patients with seven cancer types (17 series, 47 scans). \textbf{TumorMap-Lung}~\cite{tumormap}: chest CT of 33 lung cancer patients before and during immunotherapy, every
lung tumor $\ge$1\,cm contoured and linked across scans.
\textbf{PanTrack}~\cite{longitrack}: 15 patients (52 visits) with lesions linked over up to 10 visits.
\textbf{NCI-MATCH}~\cite{eay131,eay131ann}: CT of 11 patients (23 scans) from the longitudinal subset of the
NCI-MATCH trial~\cite{ncimatch}, with lesions contoured and linked across scans.
\textbf{DeepLesion DLS}~\cite{dlt}: 960 directed lesion pairs with radiologist RECIST lines at both visits.
\textbf{LIDC-NLST501}~\cite{lidcnlst}: screening CTs, each read independently by four of 20 radiologists.

\paragraph{Settings and baselines.} In the \emph{interactive} setting, every method gets one click per reference
baseline lesion, and in the \emph{automatic} setting only the images. Interactive baselines are the longitudinal models
LesionLocator~\cite{lesionlocator} and LongiTrack~\cite{longitrack}, and registration followed by the universal lesion
segmenter ULS23~\cite{uls23}. The automatic \emph{pipeline} chains the agent's own tools by fixed rules (VoxTell
detects, uniGradICON and LongiTrack track, every candidate is accepted). The \emph{one-shot LLMs} (Qwen3.8~\cite{qwen38},
MedGemma~1.5~\cite{medgemma15}, Opus~5.5~\cite{opus55}) get all visits in one prompt and answer once
(\cref{sec:supp-vlm}). \emph{Claude Code} runs the agent's own LLM (Claude Opus~5.5) with a shell and scientific Python
on the same images and task. Unlike \ours, which can only call its tools, it may write and run any code. To isolate
the harness, we also give it the five 3D models of \ours as Python functions, but not the skill
(\cref{tab:ablations,sec:supp-coding}). \ours uses the same skill, tools and settings on every dataset. All methods are scored by the same evaluation protocol (\cref{sec:supp-eval}).

\paragraph{Metrics.} \emph{F1}: baseline detection of the measurable reference lesions ($\ge$0.5\,ml), matched by mask
overlap; \emph{Rec.}: lesion recall on PanTrack and NCI-MATCH, where only some lesions are annotated (PanTrack: the pancreatic
tumor) and the response cannot be scored; \emph{Track}: share of linked reference lesions found again at the next visit (a lesion never found counts as
a miss); \emph{LD}: median absolute long-axis error of the targets against the clinical measurement or reference mask; \emph{Resp.}:
balanced accuracy of the overall response at each follow-up against the reference (\cref{sec:supp-eval}).

\paragraph{Implementation.} The tools run on one 24\,GB GPU. The agent uses Claude Opus~5.5 through the Claude Agent
SDK, or Qwen3.8-27B through the same harness. \Cref{tab:main} reports one read per study. To measure run-to-run variability, we read TumorMap, the largest dataset in \cref{tab:main}, three times (\cref{tab:supp-runs}). No clinician answers \tool{ask\_clinician} during evaluation (\cref{sec:supp-app}), so the agent decides itself.

\begin{figure*}[t]
\centering
\includegraphics[width=\linewidth]{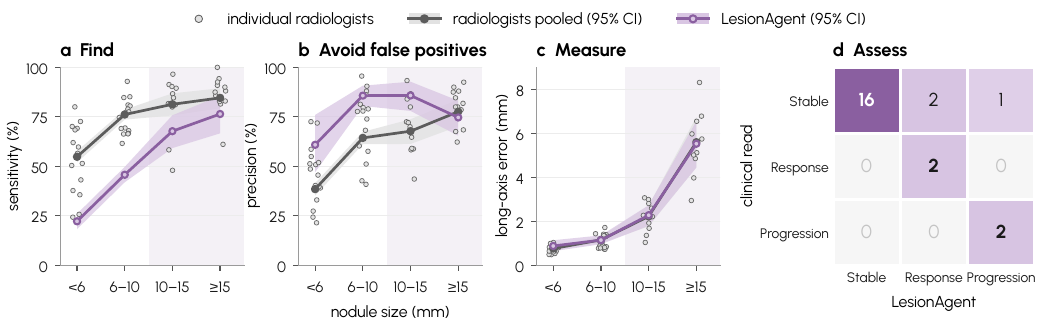}
\caption{\textbf{\ours next to human readers in detection, measurement and response assessment.} (a--c)~On LIDC,
\ours reads 436 screening CTs as a fifth radiologist, scored like each of the four readers against the nodules marked
by at least two of the others. It finds fewer nodules, above all small ones (a), but its findings are more often real
(b) and it measures them as accurately as the radiologists (c). Shading marks the nodules measurable under RECIST ($\ge$10\,mm). (d)~On the 23 HCUCH follow-ups, its response matches
the clinical reference 20 times, missing no response and no progression.}
\label{fig:humans}
\end{figure*}

\section{Results and Discussion}
\label{sec:results}

\paragraph{\ours is the strongest method end to end.} Given only the images, \ours tracks more lesions and measures
them more accurately than every automatic baseline on all four datasets (\cref{tab:main}). On HCUCH, its detection F1 is 16 points above that of the pipeline built from the same 3D models,
at less than half its measurement error. The gains hold across repeated reads, as even the weakest of three TumorMap
reads beats every other automatic method in detection, tracking and response (\cref{tab:supp-runs}). With one click
per lesion, \ours tracks every clinical target on HCUCH and 97\% of the NCI-MATCH lesions, where the baselines track
at most 84\%. Most of the gap between interactive and automatic reads lies in detection, since a lesion that is never found cannot be tracked. Built
only from open-weight models, \ours\ (Qwen3.8) is still the best open method on most metrics, above all the response.

\paragraph{Fixed rules turn model errors into false progressions.} The pipeline runs the same 3D models as \ours, but by fixed rules. Over the 56 follow-ups of HCUCH and TumorMap, it calls progression falsely 33 times, \ours twice (\cref{fig:confusion}). A fixed rule cannot tell a new lesion from a baseline lesion that detection missed, whereas the agent compares every new finding with the baseline before it records a new lesion.

\paragraph{The agent catches what the models get wrong.} \Cref{fig:traces} shows one real case per step of a read.
Sweeping the lungs slab by slab, the agent adds a nodule that no detector proposed. A second opinion exposes a VoxTell
mask that includes a vessel, and the agent's choice cuts the long-axis error from 9.3 to 1.4\,mm. When registration
and LongiTrack land on a neighbouring structure without any warning, the agent notices the different anatomy and finds
the lesion 30\,mm away. A small new spot is not called progression until it has grown at the next scan. The guideline
engine then derives the expert's response at both follow-ups. The agent also knows where it is unsure. For four of its 11 wrong responses on HCUCH and TumorMap, it had asked the clinician about exactly the deciding point. No clinician answers in our automatic evaluation, but in practice one answer would settle such a call (\cref{sec:supp-app}).

\begin{table*}[t]
\begin{minipage}[t]{\columnwidth}\centering
\caption{\textbf{Ablations on HCUCH} (automatic). Top: Each row removes or replaces one part of \ours (Claude Opus~5.5).
Without the 3D models, detection halves, and without the guideline engine, response accuracy drops. Without
self-verification (no renders, no checks in the tools), every metric drops. Bottom: Claude Code and fixed rules run the
same 3D models without the harness, and Claude Code finds far fewer lesions. Tools and \$ are the median tool calls and
API cost per read ($^\dagger$own GPU, estimated at the lowest public API price).}
\label{tab:ablations}
\small
\setlength{\tabcolsep}{2.6pt}
\renewcommand{\arraystretch}{1.08}
\begin{tabular}{@{}lcccccc@{}}
\toprule
 & F1\,$\uparrow$ & Track\,$\uparrow$ & LD\,$\downarrow$ & Resp.\,$\uparrow$ & Tools & \$ \\
\midrule
\rowcolor{gray!15}\ours & \textbf{78} & \textbf{88} & 1.9 & \textbf{95} & 101 & 2.7 \\
\quad -- guideline engine & \multicolumn{3}{c}{\color{gray}\textit{same lesions}} & 79 & 101 & 2.7 \\
\quad -- skill & 68 & 80 & 2.6 & \textbf{95} & 60 & 1.3 \\
\quad -- self-verification & 76 & 86 & 2.5 & 91 & 42 & 0.8 \\
\quad -- all 3D models & 36 & 61 & 2.0 & 61 & 65 & 1.9 \\
\quad with Qwen3.8 & 73 & 82 & 4.4 & 62 & 169 & 0.4$^\dagger$ \\
\midrule
\multicolumn{7}{@{}l}{\textit{Same 3D models, no harness}} \\
\quad Claude Code & 41 & 86 & \textbf{1.6} & 93 & 61 & 1.6 \\
\quad Pipeline (fixed rules) & 62 & 86 & 4.5 & 40 & -- & 0 \\
\bottomrule
\end{tabular}

\end{minipage}\hfill
\begin{minipage}[t]{\columnwidth}\centering
\caption{\textbf{Zero-shot tracking on the DeepLesion benchmark}~\cite{dlt}. Given a lesion at one visit, each method
is tasked to retrieve it at the other. {\setlength{\fboxsep}{1pt}\colorbox{gray!15}{Grey models}} were trained on the benchmark. Without training on its lesion pairs, \ours retrieves more lesions than every trained tracker and measures them within about 2\,mm of the radiologist. CPM@10: share of lesions retrieved within 10\,mm, MED: mean centre error (mm), Empty: share of
empty follow-up masks, LD: median long-axis error of the retrieved lesions against the radiologist (mm). $^\dagger$Published numbers.}
\label{tab:tracking}
\small
\setlength{\tabcolsep}{3.5pt}
\resizebox{\linewidth}{!}{%
\begin{tabular}{@{}llcccc@{}}
\toprule
Method & Venue & CPM@10\,$\uparrow$ & MED\,$\downarrow$ & Empty\,$\downarrow$ & LD$\downarrow$ \\
\midrule
\multicolumn{6}{@{}l}{\textit{Point trackers}} \\
Affine$^\dagger$~\cite{dlt} & -- & 0.483 & 11.2 & -- & -- \\
\rowcolor{gray!15} VoxelMorph$^\dagger$~\cite{voxelmorph} & TMI'19 & 0.499 & 10.9 & -- & -- \\
DEEDS$^\dagger$~\cite{deeds} & TMI'13 & 0.719 & 7.4 & -- & -- \\
\rowcolor{gray!15} DLT$^\dagger$~\cite{dlt} & CVPR'21 & 0.789 & 7.0 & -- & -- \\
\rowcolor{gray!15} TransT$^\dagger$~\cite{transt} & CVPR'21 & 0.796 & 7.6 & -- & -- \\
\rowcolor{gray!15} TLT$^\dagger$~\cite{tlt} & MICCAI'22 & 0.874 & 6.0 & -- & -- \\
\midrule
\multicolumn{6}{@{}l}{\textit{Trackers that also segment}} \\
LesionLocator~\cite{lesionlocator} & CVPR'25 & 0.556 & 14.5 & 37\% & 4.0 \\
LongiTrack~\cite{longitrack} & MICCAI'26 & 0.796 & 6.2 & 20\% & 9.1 \\
\multicolumn{2}{@{}l}{\textbf{\ours} (zero-shot)} & \textbf{0.890} & \textbf{4.4} & \textbf{0.2\%} & \textbf{2.0} \\
\bottomrule
\end{tabular}}

\end{minipage}
\end{table*}

\paragraph{The harness, not the LLM, makes the difference.} Without 3D models, LLMs do not see the lesions. Given all visits in one prompt, Opus~5.5 finds only 13\% of the HCUCH lesions, and Qwen3.8 and MedGemma almost none.
Opus still names the response well on HCUCH, where most follow-ups are stable, but its responses rest on lesions it
never found, whereas every response of \ours rests on lesions a clinician can check. Claude Code runs the same LLM and
may write any code, yet reaches a detection F1 of only 31\%. Given the same 3D models as \ours, it tracks and measures
as well but still finds far fewer lesions (F1 41 vs.\ 78\%, \cref{tab:ablations}). In a patient with many lung metastases, it notes them but measures only
the two largest liver and six lung lesions, enough for the response. In \ours, every detector candidate stays open in
the registry until the agent accepts or rejects it. Inside the harness, even the open-weight Qwen3.8 rises from 6 to 73\% F1.

\paragraph{Every part of the harness contributes.} In the ablations on HCUCH (\cref{tab:ablations}), removing the 3D models halves detection, and replacing the LLM by
fixed rules collapses the response. Without the skill, the agent leaves many detector candidates
undecided and finds, tracks and measures worse, but still finds more lesions than Claude Code with the same 3D models.
Without self-verification, every metric drops. When the LLM assigns the response itself instead of the guideline
engine, agreement drops from 95 to 79\%, as it counts small spots seen for the first time as new lesions and picks
other targets.

\paragraph{Zero-shot tracking beats trained trackers.} To evaluate tracking alone, we use the DeepLesion benchmark
(\cref{tab:tracking}). Without training on it, \ours retrieves 89.0\% of the lesions, ahead of TLT (87.4\%), the best
of the trackers trained on its lesion pairs, and locates them more precisely (mean centre error 4.4 vs.\ 6.0\,mm). These trackers return only a point. Of the trackers that also segment,
LongiTrack and LesionLocator return an empty mask for 20 and 37\% of the pairs. \ours almost never does and measures the long axis within a median of 2.0\,mm of the radiologist. \ours uses the
same models as LongiTrack, so the gain comes from what it does when they fail. It compares both visits around the
propagated point, searches the overlapping slices and re-segments from a corrected point.

\paragraph{\ours measures like a radiologist but finds fewer lesions.} To compare \ours with radiologists, we add it to LIDC as a fifth reader and score it like
each radiologist against the others. It measures nodules of at least 10\,mm as accurately as the radiologists and
reports fewer false positives, but finds fewer nodules (sensitivity 67 vs.\ 83\%, \cref{fig:humans}). On
clinical follow-up CT (HCUCH), it agrees with the radiologist's response at 20 of 23 follow-ups, including all four
changes in response, and its time to progression matches in 11 of 12 series. It chooses 17 of the radiologist's 25 target lesions (\cref{sec:supp-eval}). For reference, in other cohorts, two
radiologists choose different targets in most patients and agree on the response in 74\% of
cases~\cite{kuhl2019recist,karmakar2019}.

\paragraph{Limitations and outlook.} \ours is only as strong as the models it operates. Small lesions, its main gap
to radiologists, depend on the detector, and with an open-weight LLM the response still trails the best closed model.
Because the harness is model-agnostic, every better detector or open model plugs in without retraining. Public longitudinal cohorts with clinical reads are still small, and HCUCH records only the radiologists' target lesions, not their non-target or new-lesion calls. Larger cohorts will test more responses and progressions, and a reader study will measure what clinicians add through \tool{ask\_clinician}.

\section{Conclusion}
\label{sec:conclusion}

We presented \ours, an LLM agent that monitors tumors across longitudinal CT by composing frozen 3D foundation models,
remembering every lesion across visits, verifying the models' outputs in the loop and leaving the final assessment to a
deterministic guideline engine. Without task-specific training, it outperforms all automatic baselines end to end,
agrees with the radiologists' response at 20 of 23 clinical follow-ups and retrieves more DeepLesion lesions than every
tracker trained on that benchmark. Its harness, not its LLM, makes the difference. Given the same LLM and 3D models,
Claude Code reaches a detection F1 of 41\% and \ours 78\%, and the model-agnostic harness lifts an open-weight LLM from
6 to 73\%. Every response it reports is grounded in lesions a clinician can check.

\section*{LLM Usage Statement}
LLMs are naturally part of the method and of the baselines as this paper is in the realm of agentic AI: \ours runs on Claude Opus~5.5 (Claude Agent SDK) or Qwen3.8-27B,
Claude Code runs Claude Opus~5.5, and the one-shot baselines are Qwen3.8-27B, MedGemma~1.5 4B, MedGemma 27B and Claude Opus~5.5
(\cref{sec:experiments}). We also used Claude Code (Claude Opus~5.5) as an assistant to help implement, debug and run code,
including the evaluation and plotting scripts, to double-check numbers, and to help draft and edit the text. The authors checked
all code, results and text and take full responsibility for the content.

\section*{Acknowledgments}
This work was supported by the NCT Heidelberg Image Data Analysis Infrastructure (IDAI). M.R.\ was supported by the
Google PhD Fellowship Program. B.H.\ was funded by the Vector Stiftung. Y.K.\ was funded by the Deutsche
Forschungs\-gemeinschaft (DFG, German Research Foundation) -- 402688427. Y.K.\ was supported by the Helmholtz
Association under the joint research program ``HIDSS4Health -- Helmholtz Information and Data Science School for Health''. This research was supported by the Arc Institute Graduate Fellowship to A.L.

\newpage

{
    \small
    \bibliographystyle{ieeenat_fullname}
    \bibliography{main}
}

\clearpage
\maketitlesupplementary
\appendix
\setcounter{figure}{0}\setcounter{table}{0}
\renewcommand{\thefigure}{S\arabic{figure}}\renewcommand{\theHfigure}{S\arabic{figure}}
\renewcommand{\thetable}{S\arabic{table}}\renewcommand{\theHtable}{S\arabic{table}}

\newcommand{\suppcontents}[2]{\par\noindent\parbox[t]{\dimexpr\linewidth-1.6em\relax}{\hangindent=1.4em\hangafter=1
  \hyperref[#1]{\makebox[1.4em][l]{\ref*{#1}.}#2}\dotfill}\hfill\makebox[1.6em][r]{\pageref*{#1}}\par}
\noindent\textbf{Contents}\par\smallskip
\begingroup\hypersetup{linkcolor=black}
\suppcontents{sec:supp-related}{Related Methods at a Glance}
\suppcontents{sec:supp-data}{Data}
\suppcontents{sec:supp-eval}{Evaluation Protocol}
\suppcontents{sec:supp-tools}{Tool Reference}
\suppcontents{sec:supp-traces}{More Agent Reads}
\suppcontents{sec:supp-app}{Interactive Application}
\suppcontents{sec:supp-vlm}{One-Shot LLM Baselines}
\suppcontents{sec:supp-coding}{Claude Code Baseline}
\suppcontents{sec:supp-skill}{The Skill}
\endgroup\medskip

\section{Related Methods at a Glance}
\label{sec:supp-related}

\ours is, to our knowledge, the first method that performs the whole read on longitudinal CT end to end: from the
images alone it finds the lesions, follows them through every visit, measures them, decides whether a follow-up
finding is new and assigns the response. \Cref{tab:supp-related} shows that prior work solves parts of this.
\begin{itemize}
\item \textbf{Foundation models} solve one perception step each and must be told where to look or what to
  segment: a text prompt (VoxTell), a click (nnInteractive, ULS23), a fixed task (TotalSegmentator) or two scans to
  align (uniGradICON).
\item \textbf{Lesion trackers} locate a lesion that a reader has found in a follow-up scan~\cite{hering2021wholebody,lesionlocator,longitrack}.
  A wrong match stays wrong, and none decides whether a finding is new.
\item \textbf{RECIST automation} measures reader-selected lesions~\cite{dahm2024automated} or classifies the response
  from the radiologist's report~\cite{mergen2026llm}. Our Registration$\to$ULS baseline follows the first recipe with
  public models.
\item \textbf{Medical agents} that operate 3D models answer one query about one study~\cite{ctagent,radagent,3dmedagent}.
  The agents for follow-up imaging assess brain tumors on MRI, letting the language model assign the
  category~\cite{erdur2026agentic} or applying fixed rules to volumes from a separate segmentation
  network~\cite{jabal2026btrads}. We therefore
  compare with Claude Code running the same LLM (\cref{sec:supp-coding}).
\end{itemize}
\ours operates VoxTell, nnInteractive, uniGradICON, LongiTrack and TotalSegmentator as tools (T in \cref{tab:supp-related}) and adds
what they lack: a search for missed lesions, verification of every model output it relies on, a comparison of every
candidate new lesion with the baseline, and a deterministic guideline engine for every number and category.

\begin{table*}[t]
\centering
\caption{\textbf{What related methods cover of tumor monitoring.} Each prior method solves part of the read, and only
\ours covers every step. A \cmark\ marks a step the method does itself and $\circ$ a step it does in part, \eg only for
target lesions, only in 2D or with code but no weights. \emph{Find} means locating lesions without being told where
they are, and \emph{Correct} detecting and correcting wrong model outputs. The last column gives how we use each
method: as a tool of \ours (T), as a baseline (B, $^\dagger$numbers from its paper) or by the same recipe with public
models (R).}
\label{tab:supp-related}
\small
\setlength{\tabcolsep}{3.2pt}
\resizebox{\textwidth}{!}{%
\begin{tabular}{@{}lllccccccccccc@{}}
\toprule
 & Method & Venue & 3D & \shortstack{Several\\scans} & \shortstack{Find\\lesions} & Measure & Track &
 \shortstack{Correct\\errors} & \shortstack{New\\lesions} & Response & Agent & Code & Here \\
\midrule
\multirow{5}{*}{\shortstack[l]{3D foundation\\models}}
 & VoxTell~\cite{voxtell} & CVPR'26 & \cmark & -- & \cmark & \cmark & -- & -- & -- & -- & -- & \cmark & T, B \\
 & nnInteractive~\cite{nninteractive} & arXiv'25 & \cmark & -- & -- & \cmark & -- & -- & -- & -- & -- & \cmark & T \\
 & ULS23~\cite{uls23} & MedIA'25 & \cmark & -- & -- & \cmark & -- & -- & -- & -- & -- & \cmark & B \\
 & uniGradICON~\cite{unigradicon} & MICCAI'24 & \cmark & \cmark & -- & -- & $\circ$ & -- & -- & -- & -- & \cmark & T, B \\
 & TotalSegmentator~\cite{totalsegmentator} & Radiol.\ AI'23 & \cmark & -- & $\circ$ & $\circ$ & -- & -- & -- & -- & -- & \cmark & T \\
\midrule
\multirow{4}{*}{\shortstack[l]{Lesion\\trackers}}
 & DLT~\cite{dlt} & CVPR'21 & \cmark & \cmark & -- & -- & \cmark & -- & -- & -- & -- & -- & B$^\dagger$ \\
 & TLT~\cite{tlt} & MICCAI'22 & \cmark & \cmark & -- & -- & \cmark & -- & -- & -- & -- & $\circ$ & B$^\dagger$ \\
 & LesionLocator~\cite{lesionlocator} & CVPR'25 & \cmark & \cmark & -- & \cmark & \cmark & -- & -- & -- & -- & \cmark & B \\
 & LongiTrack~\cite{longitrack} & MICCAI'26 & \cmark & \cmark & -- & \cmark & \cmark & $\circ$ & -- & -- & -- & \cmark & T, B \\
\midrule
\multirow{3}{*}{\shortstack[l]{RECIST\\automation}}
 & Semi-auto.\ RECIST~\cite{tang2018recist} & MICCAI'18 & $\circ$ & -- & -- & \cmark & -- & -- & -- & -- & -- & -- & -- \\
 & Automated RECIST~\cite{dahm2024automated} & Cancers'24 & \cmark & \cmark & -- & \cmark & \cmark & -- & -- & $\circ$ & -- & $\circ$ & R \\
 & LLMs on reports~\cite{mergen2026llm} & Sci.\ Rep.'26 & -- & -- & -- & -- & -- & -- & -- & \cmark & -- & \cmark & -- \\
\midrule
\multirow{3}{*}{Agents}
 & CT agents~\cite{ctagent,radagent,3dmedagent} & arXiv'25--26 & \cmark & -- & $\circ$ & $\circ$ & -- & -- & -- & -- & \cmark & $\circ$ & -- \\
 & Brain MRI agents~\cite{erdur2026agentic,jabal2026btrads} & arXiv'26 & \cmark & \cmark & $\circ$ & \cmark & $\circ$ & -- & $\circ$ & \cmark & \cmark & $\circ$ & -- \\
 & Claude Code (Opus~5.5) & -- & \cmark & \cmark & \cmark & \cmark & \cmark & -- & \cmark & \cmark & \cmark & -- & B \\
\midrule
 & \textbf{\ours} & -- & \cmark & \cmark & \cmark & \cmark & \cmark & \cmark & \cmark & \cmark & \cmark & \cmark & \\
\bottomrule
\end{tabular}}
\end{table*}

\section{Data}
\label{sec:supp-data}

\begin{itemize}
\item \textbf{HCUCH}~\cite{hcuch} (CC~BY~4.0): clinical RECIST~1.1 target measurements by the radiologist.
  We use the 14 patients (17 thorax or abdomen series, 47 scans) with at least one follow-up. 23 follow-ups have
  clinical target measurements and form the response reference.
\item \textbf{TumorMap-Lung}~\cite{tumormap}: the 33 patients with metastatic non-small cell lung cancer and at least
  one follow-up, chest CT before and after the start of immunotherapy, with every lung tumor $\ge$1\,cm contoured by two
  radiation oncologists and labelled consistently across scans.
\item \textbf{PanTrack}~\cite{longitrack}: the 15 patients (52 visits) of the official validation split, lesions
  linked across up to 10 visits. Only the pancreatic tumor is annotated in every scan (liver metastases partly,
  lymph nodes rarely), so detection is scored as the recall of the pancreatic tumor and the response is not scored.
\item \textbf{NCI-MATCH}~\cite{eay131,eay131ann} (CC~BY~4.0): CT from the NCI-MATCH trial~\cite{ncimatch} with
  expert contours of at most five lesions per scan and of some new lesions, linked across
  scans. Of the 18 patients with contoured CT at two or more visits, we use the 11 whose scans cover all contoured lesions at every visit. We checked every link across scans visually and corrected the wrong ones.
  Due to the partial annotation, detection is scored as recall and the response is not scored.
\item \textbf{DeepLesion DLS}~\cite{dlt}: 480 lesion pairs (960 directed) from 105 patients, each with the
  radiologist's RECIST long and short axis at both visits, in sub-volumes of $\sim$80\,mm along $z$. These are the official test pairs of the benchmark, on which every method in \cref{tab:tracking} is evaluated.
\item \textbf{LIDC-NLST501}~\cite{lidcnlst}: the 436 of the 501 NLST screening CTs whose patients have a
  further NLST screening CT on TCIA, each read independently by four of 20 radiologists. We use the scans as single time points for detection and measurement against human readers.
\end{itemize}
\paragraph{Response references.} Both references apply the RECIST~1.1 rules to expert annotations. \emph{HCUCH}: the
radiologist's own target lesions and their diameters from the clinical read, per CT series. The reference is the
response of these targets (the dataset records no non-target or new-lesion assessment). Thorax and abdomen series of a
visit are separate cases because the clinical targets are given per series. \emph{TumorMap}: the expert tumor
segmentations, linked across scans. The targets are chosen by the RECIST rules (the largest measurable baseline tumors, at most two per
organ and five in total) and measured on the masks. Tumors that appear only at follow-up are new lesions, and the other baseline tumors are non-targets. None of the four longitudinal datasets is in the training data of the agent's tools (LongiTrack~v1.0 is trained only on
AutoPET~IV).

\paragraph{Data use and ethics.} All datasets are de-identified and were released by their owners, who obtained the
ethics approvals and patient consent, or a waiver of consent, for the original studies. We use them under their licences or data-use agreements, and the owners of
TumorMap-Lung confirmed in writing that our use is permitted. No new patient data were collected for this work.
\ours is a research system and not validated for clinical use. It can miss small lesions, and a complete-looking
report can invite over-reliance, so every response is meant to be checked by a radiologist. The reading application
supports this by showing every lesion, measurement and tool call behind the response (\cref{sec:supp-app}).

\section{Evaluation Protocol}
\label{sec:supp-eval}

\paragraph{Interactive baselines.} Every interactive method gets one click per reference baseline lesion and
uses only its own models. LesionLocator and LongiTrack segment the clicked lesion at baseline themselves and track
every follow-up from the baseline (LongiTrack with the click propagated by uniGradICON). Registration$\to$ULS
propagates the click with uniGradICON and segments the lesion at every visit with the ULS23 model.

\paragraph{Fixed rules.} The pipeline and the interactive baselines replace the reader's judgement by fixed rules: an
empty follow-up mask means the lesion has disappeared, and the targets are the best valid set of the same ranking the
agent sees (\cref{sec:selection}). Only the pipeline searches for new lesions: a follow-up VoxTell candidate
$\ge$10\,mm and $\ge$20\,mm away from every baseline finding is a new lesion. The interactive baselines follow the
clicked lesions only and report no new lesions.

\paragraph{Metrics.} \emph{F1}: a reference lesion counts as found if a lesion reported by the method overlaps it. For
methods that only measure, the overlap is taken with their calipers (the long- and short-axis lines of a RECIST
measurement): a RECIST line drawn across a lesion always overlaps it. Lesions the agent rejected do not count. Detector candidates it left undecided, almost only in the ablation
without the skill, have no mask of their own. They count like reported lesions, as found if their centre lies inside a
reference lesion and as a false positive otherwise. Recall is computed over measurable
reference lesions ($\ge$0.5\,ml), precision over all found lesions. PanTrack and NCI-MATCH contour only some lesions
per patient. A found lesion outside them can be neither confirmed nor counted as a false positive, so we report
the recall alone. TumorMap contours only lung tumors $\ge$1\,cm,
so there a found lesion that matches no reference lesion counts as a false positive only from 10\,mm long axis, for
every method. \emph{Track}: the share of linked reference lesions that the method's lesion still overlaps at the
follow-up. Lesions never found count as failures. \emph{LD}: median absolute long-axis error of the targets
against the clinical measurement (TumorMap: against the long axis of the matched reference mask at every visit where the lesion is annotated). \emph{Resp.}: balanced accuracy of the per-visit overall response against the clinical reference (HCUCH)
or the mask-derived reference (TumorMap): the agreement is computed per reference category and
averaged, so a method that always answers stable disease scores 33\%. A visit without a parsable answer counts as wrong. NCI-MATCH is
scored like PanTrack, on all contoured lesions (lymph nodes: short axis). In \cref{tab:main}, $\pm$ is the standard deviation over 1000 bootstrap resamples of the lesions (Resp.: of the follow-ups).

\paragraph{DeepLesion.} We follow the DLT protocol on all 960 directed pairs: given a lesion's centre in one
sub-volume, predict its centre in the other. We score with the evaluation code released with DLT: CPM@10 counts
predictions within $\min(r, 10\,\mathrm{mm})$ of the reference centre ($r$: lesion radius), and MED is the mean centre
error.
For methods that segment, the centre is that of the predicted component (an empty mask falls
back to the method's starting point). Empty is the share of pairs with an empty follow-up mask, and LD the median
long-axis error against the radiologist on correctly retrieved lesions. Released methods are run as released, and the agent is run once per pair. No tool was trained
to track lesions on DeepLesion: the agent tracks with uniGradICON and LongiTrack (AutoPET~IV only).

\paragraph{LIDC fifth reader.} For each radiologist we score detection against the nodules marked by at least two of
the other three readers, and the agent against each of these four references (averaged), so that both face the same
references. Measurement compares each
long axis with the mean of the other readers. The agent read all 436 scans (148 with a nodule $\ge$10\,mm). Reader values
are pooled over all reads of the same scans, with 95\% CIs from a bootstrap over scans. For nodules $\ge$10\,mm, \ours
reaches a sensitivity of 67\% [60, 74] against 83\% [78, 87] for the readers, with 0.07 [0.05, 0.10] against
0.14 [0.12, 0.18] false positives per scan and a long-axis error of 4.2 [3.5, 4.9] against 4.1 [3.4, 4.9]\,mm.

\begin{figure}[t]
\centering
\includegraphics[width=\linewidth]{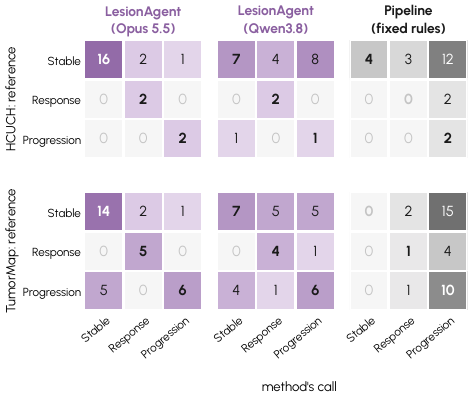}
\caption{\textbf{The agent's judgement removes false progressions.} Response at every follow-up with a reference
(rows) against the call (columns), automatic setting: HCUCH (23 follow-ups, clinical measurements) and TumorMap (33). Response: complete or
partial, merged here (\cref{tab:main} scores them apart). With the same 3D models, the fixed pipeline calls progression falsely at 33 of the 56 follow-ups, \ours
only twice.}
\label{fig:confusion}
\end{figure}

\paragraph{Statistics.} \Cref{tab:supp-stats} compares \ours with the strongest baselines at the patient level, and
\cref{fig:confusion} shows where the response calls of \ours and of the pipeline go.

\begin{table}[t]
\centering
\caption{\textbf{\ours's gains hold at the patient level.} \ours $-$ baseline (in points, LD in mm, negative is better)
with 95\% CI from 4000 paired bootstrap resamples of the patients (HCUCH: series). Shaded and bold: the CI excludes
zero. \ours tracks lesions significantly better than Claude Code and Opus one-shot on both datasets and
assesses the response significantly better than the pipeline and the open-weight agent. On HCUCH, it also finds and
measures lesions significantly better than Opus one-shot.}
\label{tab:supp-stats}
\footnotesize
\providecommand{\dd}[3]{\begin{tabular}[c]{@{}c@{}}#1\\[-2.5pt]{\tiny[#2,\,#3]}\end{tabular}}
\providecommand{\ds}[3]{\cellcolor{green!12}\begin{tabular}[c]{@{}c@{}}\textbf{#1}\\[-2.5pt]{\tiny[#2,\,#3]}\end{tabular}}
\setlength{\tabcolsep}{2.2pt}
\resizebox{\linewidth}{!}{%
\begin{tabular}{@{}lccccccc@{}}
\toprule
 & \multicolumn{4}{c}{HCUCH} & \multicolumn{3}{c}{TumorMap} \\
\cmidrule(lr){2-5}\cmidrule(l){6-8}
\ours vs. & F1 & Track & LD & Resp. & F1 & Track & Resp. \\
\midrule
Pipeline & \ds{+16}{5}{24} & \dd{+2}{$-$11}{18} & \ds{$-$2.6}{$-$5.3}{$-$0.5} & \ds{+54}{25}{86} & \ds{+10}{4}{16} & \ds{+21}{13}{33} & \ds{+42}{22}{57} \\
Claude Code & \ds{+47}{4}{68} & \ds{+25}{9}{43} & \dd{$-$0.4}{$-$2.0}{1.6} & \dd{+15}{$-$8}{23} & \dd{+13}{$-$12}{28} & \ds{+38}{15}{49} & \dd{+5}{$-$11}{23} \\
Qwen3.8 agent & \dd{+5}{$-$7}{16} & \dd{+6}{0}{21} & \ds{$-$2.4}{$-$7.9}{$-$0.4} & \ds{+33}{11}{77} & \ds{+12}{6}{20} & \ds{+18}{9}{32} & \ds{+27}{6}{49} \\
Opus one-shot & \ds{+57}{17}{74} & \ds{+35}{16}{53} & \ds{$-$4.3}{$-$12.9}{$-$1.6} & \dd{+7}{$-$2}{24} & \dd{+23}{$-$5}{39} & \ds{+48}{28}{56} & \dd{+10}{$-$10}{32} \\
\bottomrule
\end{tabular}}
\end{table}

\paragraph{Run-to-run variability.} A read of the agent is stochastic: single decisions, such as the choice of
targets or whether to accept a new lesion, can change the response. To measure this, we read all 33 TumorMap
studies, the largest of the four longitudinal datasets, three times with the same code and settings
(\cref{tab:supp-runs}). Read~1, the first, is the one in \cref{tab:main}. F1 and response vary by 1.5 points (SD)
across reads and tracking by 3.5. Of the 33 follow-ups, 26 receive the same response call in all three reads
(pairwise Cohen's $\kappa$ 0.76--0.80). Six follow-ups are called wrong in every read, so most wrong calls are
systematic rather than chance. Against the other automatic methods, even the weakest read is above all of them in
F1, tracking and response, and the long-axis error of the three reads is on par with Claude Code (4.0--4.5
vs.\ 4.4\,mm). Every significant difference on TumorMap in \cref{tab:supp-stats} holds for each of the three reads.
We did not repeat the other agentic baselines or Opus one-shot for cost
reasons, while the remaining baselines involve no sampling.

\begin{table}[t]
\centering
\caption{\textbf{Run-to-run variability on TumorMap.} Three reads of all 33 studies by \ours\ (Opus~5.5, automatic)
with the same code and settings. Read~1 is the one in \cref{tab:main}. Below, the strongest other automatic methods of
\cref{tab:main}, read once. F1, Track and Resp.\ in \%, LD in mm.}
\label{tab:supp-runs}
\footnotesize
\setlength{\tabcolsep}{3.6pt}
\begin{tabular}{@{}lcccc@{}}
\toprule
 & F1 & Track & LD\,$\downarrow$ & Resp. \\
\midrule
\ours, read 1 & 62.3 & 75.4 & 4.0 & 79.0 \\
\ours, read 2 & 61.8 & 73.7 & 4.5 & 76.1 \\
\ours, read 3 & 59.6 & 68.6 & 4.4 & 77.0 \\
Mean $\pm$ SD & 61.2\,$\pm$\,1.5 & 72.6\,$\pm$\,3.5 & 4.3\,$\pm$\,0.3 & 77.4\,$\pm$\,1.5 \\
\midrule
Pipeline & 52.2 & 54.2 & 8.1 & 37.0 \\
Claude Code & 49.7 & 37.3 & 4.4 & 74.4 \\
Opus~5.5 one-shot & 39.7 & 27.1 & 8.4 & 68.7 \\
\ours\ (Qwen3.8) & 50.3 & 57.6 & 10.3 & 51.9 \\
\bottomrule
\end{tabular}
\end{table}

\paragraph{Target selection.} The pipeline, which simply takes the best set of our ranking (\cref{sec:selection}), already chooses 64\% of the radiologist's target lesions on HCUCH. \ours, which also scores how reproducibly each lesion can be measured and may justify another set, chooses 68\%. For reference, two radiologists choose the
same target lesions in only 41\% of patients~\cite{kuhl2019recist}. Identical target sets are rare here, because the
radiologist distributes up to five targets per patient across thorax and abdomen, while each series is read on its
own.

\paragraph{Wrong responses.} Most of the 11 wrong response calls (of 56 follow-ups on HCUCH and TumorMap) lie near a
response threshold or follow another target measurement than the reference. \ours misses five of the 11 TumorMap
progressions, two of them because it waits for a new lesion to be confirmed at the next scan, which TumorMap does not
have.

\section{Tool Reference}
\label{sec:supp-tools}

\Cref{tab:supp-tools} lists the 18 tools by role. They check their outputs in code (\cref{sec:verification}) with
these thresholds:
\begin{itemize}
\item \textbf{Second opinion}: an nnInteractive click at the mask's centre disagrees if the long axes differ by more than
  $\max(5\,\mathrm{mm}, 30\%)$. Clicks that leak ($>$150\,ml) and masks of more than five components are not compared.
\item \textbf{Tracking}: a mask is flagged if it is empty, grows above $3\times$ or shrinks below $0.3\times$ its earlier
  volume, misses the propagated point, splits or touches another lesion.
\item \textbf{Target choice}: the response is recomputed for all near-tied target sets (same organ coverage, score within
  15\% of the best), naming the lesions whose choice changes it.
\end{itemize}

\begin{table*}[t]
\centering
\caption{\textbf{Tool reference.} The 18 tools with their input and output (typed JSON schemas, served to the model as
an in-process MCP server). Every tool returns text, and tools marked $\ast$ also return an image. Points are voxel
indices $[z, y, x]$.}
\label{tab:supp-tools}
\footnotesize
\setlength{\tabcolsep}{4pt}
\renewcommand{\arraystretch}{1.08}
\begin{tabular}{@{}ll>{\raggedright\arraybackslash}p{0.21\linewidth}>{\raggedright\arraybackslash}p{0.465\linewidth}@{}}
\toprule
Group & Tool & Input & Output \\
\midrule
Perception & \tool{detect} & visit, text prompts (\eg ``liver metastasis'') & VoxTell candidates $c_1, c_2, \dots$ with size, organ and the prompts that found them. At a follow-up, each candidate's registered position at baseline, to tell a missed lesion from a new one \\
\cmidrule(l){2-4}
 & \tool{totalsegmentator} & visit, task & the organ group of every lesion (RECIST counts targets per organ) and node hints, or more candidates from a second detector (lung nodules, liver tumors) \\
\cmidrule(l){2-4}
 & \tool{segment}$^\ast$ & one point or a box, visit, method & a lesion's mask at a visit from nnInteractive or LongiTrack and its long and short axis. A LongiTrack mask is checked as in \tool{track\_all} \\
\cmidrule(l){2-4}
 & \tool{track\_all} & target visit & every lesion propagated by uniGradICON and segmented by LongiTrack with its earlier mask as context, each with quality flags and an nnInteractive second opinion \\
\cmidrule(l){2-4}
 & \tool{refine\_lesion} & lesion, positive / negative points or scribbles & the mask corrected by nnInteractive, starting from the current mask. \texttt{undo} restores the previous one \\
\cmidrule(l){2-4}
 & \tool{edit\_mask} & lesion, operations & deterministic edits without a model: keep the largest component, remove one, fill holes, clip along an axis, subtract a neighbour \\
\midrule
Inspection & \tool{study\_info} & -- & visits, days, geometry, progress through the reading phases and the lesion table \\
\cmidrule(l){2-4}
 & \tool{view}$^\ast$ & visit, centre, lesion or candidate, mode, window & slices, a montage, a whole-scan sweep or an overview, with absolute voxel coordinates on the axes, so that a point read off the image is a valid tool input. Views of a lesion show at least 100\,mm of context \\
\cmidrule(l){2-4}
 & \tool{view\_pair}$^\ast$ & lesion (centre per visit) & the lesion at every visit side by side, with its point, the registration point and measurement lines: the check for tracking \\
\cmidrule(l){2-4}
 & \tool{inspect} & lesion, candidate or point & numbers without an image: components, extent per slice, HU inside and around the mask, organ, quality record \\
\midrule
Registry & \tool{review\_candidates}$^\ast$ & accept (name, type, organ) or reject (reason) per candidate & accepted candidates become lesions, one per separate nodule ($\ge$0.1\,ml), each with an nnInteractive second opinion (both masks side by side if they disagree) \\
\cmidrule(l){2-4}
 & \tool{update\_lesion} & lesion, fields & name, type, organ, role (target, non-target, new, rejected) and status at a visit (present, too small = 5\,mm, absent = 0\,mm, merged) \\
\cmidrule(l){2-4}
 & \tool{set\_visit\_judgement}$^\ast$ & visit, non-target status, new lesion & the reader's judgements at a follow-up. A new lesion is recorded only after an image of the baseline at its registered position (then / now) \\
\midrule
Decision & \tool{score\_lesions} & per lesion: reproducibility score 0--1, flags & the agent's judgement of how reliably each lesion can be measured again (cystic, ill-defined, confluent, \dots) \\
\cmidrule(l){2-4}
 & \tool{propose\_targets} & -- & the eligible baseline lesions and all valid target sets ($\le$5, $\le$2 per organ, measurable), ranked by organ coverage and score-weighted size (\cref{sec:selection}) \\
\cmidrule(l){2-4}
 & \tool{set\_targets} & lesions, justification & the target set (all other lesions become non-targets), refused if it breaks a RECIST rule \\
\cmidrule(l){2-4}
 & \tool{recist\_report} & -- & the engine's result per visit: sum of diameters, its change from baseline and from the smallest sum so far, target, non-target, new-lesion and overall response, distance to the response thresholds, decision-critical visits, stability under near-tied target sets \\
\midrule
Human & \tool{ask\_clinician} & kind (confirm, choose, click), question, options & the clinician's answer or click. Without an answer, the agent decides and says so \\
\bottomrule
\end{tabular}
\end{table*}

\section{More Agent Reads}
\label{sec:supp-traces}

\begin{figure*}[t]
\centering
\includegraphics[width=\linewidth]{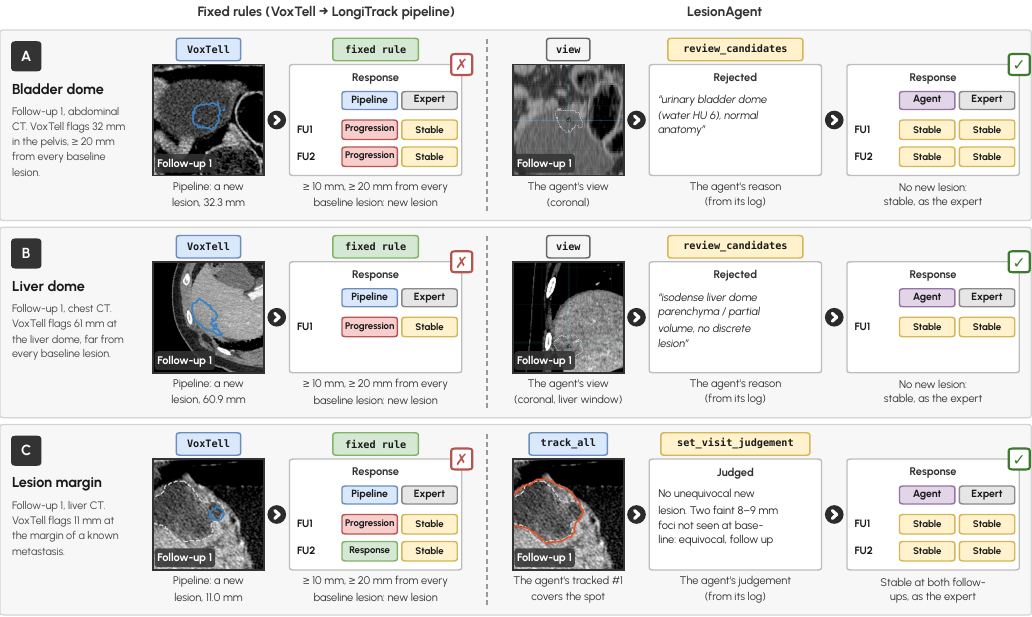}
\caption{\textbf{Fixed rules vs.\ \ours} on the same follow-up findings in three patients (HCUCH, TumorMap). The
pipeline counts every VoxTell finding of at least 10\,mm that lies at least 20\,mm from all baseline lesions as a new
lesion and calls progression. The agent looks at each finding: it rejects normal anatomy (A, B), recognises part
of a known metastasis (C) and reads stable disease as the expert does.}
\label{fig:traces-supp}
\end{figure*}

\begin{figure*}[p]
\centering
\includegraphics[width=0.97\linewidth]{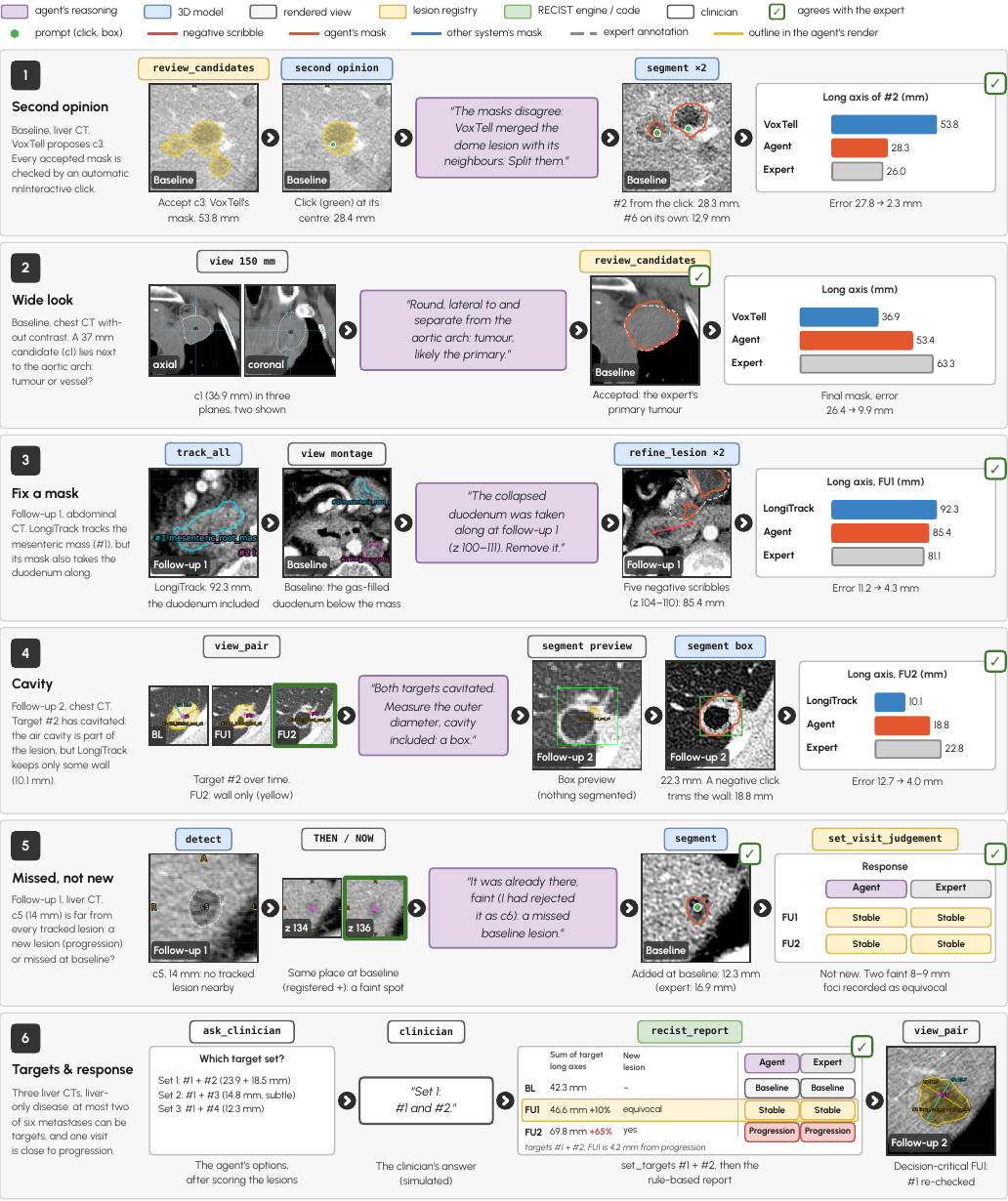}
\caption{\textbf{\ours at work, step by step.} Six reads on HCUCH and TumorMap, one per row. The left column states
the situation, and the row then follows the agent from left to right. A label above each step names it, coloured by
who or what runs it (legend). Purple boxes quote the agent's reasoning, shortened from the run log. Images with yellow or cyan
outlines are renders the agent saw, the others are CT crops with the stored masks. The clinician's answer in row~6 is
simulated for illustration. In the evaluation nobody answers, and the agent chose the same set by itself. BL, FU1,
FU2: baseline, follow-up 1, 2.}
\label{fig:traces-stack}
\end{figure*}

\Cref{fig:traces-supp} compares the fixed pipeline with the agent in three patients. \Cref{fig:traces-stack} follows
six reads of the final agent step by step, in more depth than \cref{fig:traces}: the tool calls, what the agent saw,
its notes and the result.

\section{Interactive Application}
\label{sec:supp-app}

\begin{figure*}[t]
\centering
\includegraphics[width=\linewidth]{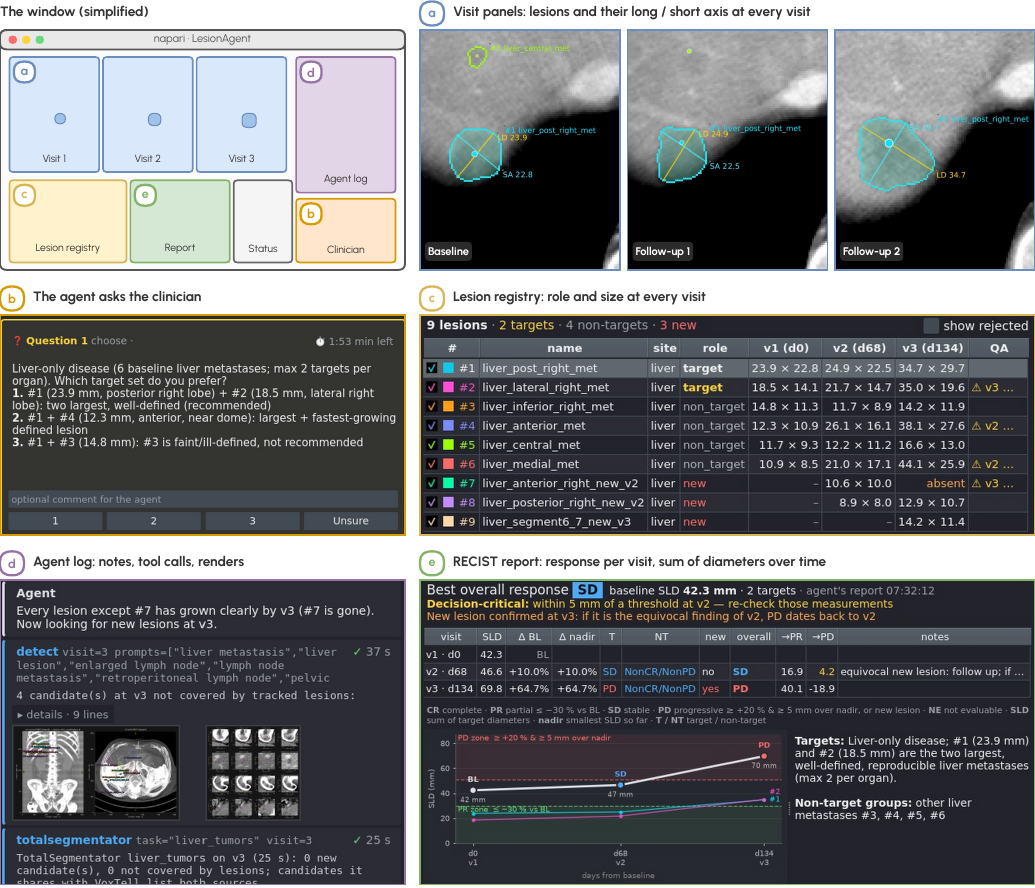}
\caption{\textbf{\ours in napari} during a live read of a patient with liver metastases (HCUCH, three visits). Top
left: the window, simplified, with the place of each zoom. (a)~Visit panels, (b)~a clinician question, (c)~the lesion
registry, (d)~the agent's log, (e)~the RECIST report from the guideline engine. The clinician sees every lesion, tool
call and number behind the response and can correct any of them.}
\label{fig:napari}
\end{figure*}

\ours ships as a napari~\cite{napari} plugin that turns the read into a reading application (\cref{fig:napari}). A launcher opens
a patient's study and selects the language model, including the open-weight Qwen3.8 on local hardware, so that no image
has to leave the hospital. The window shows the read live, as the agent works.

\noindent(a)~\textbf{Visit panels.} One panel per visit shows every lesion's outline, in the same colour at every
visit, with its registry id and its long and short axis. The panels are linked on the selected lesion: scrolling,
panning or zooming one visit moves the others to the same lesion.

\noindent(b)~\textbf{Clinician.} Each \tool{ask\_clinician} question appears as a card: a yes/no question, a choice
between options, or a click on the lesion in a visit panel. The read waits for the answer until a time limit, then
the agent decides itself.

\noindent(c)~\textbf{Lesion registry.} One row per lesion: its site, its role (target, non-target, new), its long
$\times$ short axis at every visit and its quality flags. Selecting a row focuses all panels on that lesion. The
clinician can change a lesion's role or status, reject it, or ask the agent to re-check it. A right-click in a visit
panel segments a lesion at that point. Every change runs through the agent's own tools, is logged and is reported to
the agent at its next turn.

\noindent(d)~\textbf{Agent log.} The agent's notes, every tool call with its result and duration, and the renders it
saw.

\noindent(e)~\textbf{RECIST report.} The guideline engine updates it from the registry after every change. For each
visit, it shows the sum of target diameters, its change from baseline and from the nadir (the smallest sum so far,
from which RECIST measures progression), the target, non-target, new-lesion and overall response, and the distance
to the partial-response and progression thresholds. A chart plots the sum over time. Notes flag decisions that need
attention, such as a sum within 5\,mm of a threshold.

A status panel shows the model that is running, the GPU memory and the tool calls and cost so far. A read can be
interrupted and resumed.

\paragraph{Questions during the evaluation.} All results in this paper are fully autonomous reads: nobody answers, and
the agent settles every question itself. It asks rarely and to the point: 19 questions in the 50 automatic reads of
HCUCH and TumorMap. Almost all concern a decision the response depends on: the choice of targets, a measurement near
a response threshold, or whether a finding is tumor or a mimic such as scar or pneumonitis. Many state what is at
stake, \eg ``\tool{targets \#1 and \#3 give SD ($-$20\%), \#1 and \#16 give PR. Which target set do you
prefer?}'' (SD: stable disease, PR: partial response). The questions find the difficult calls. At four of the
11 follow-ups where the autonomous read disagrees with the reference, the agent had asked about exactly the deciding
point. One example is a target sum 0.7\,mm below the progression threshold, with the question whether the growing
opacity is tumor or radiation pneumonitis. With a clinician in the loop, one answer settles such a call.
How much a clinician's answers improve the read is for a reader study with radiologists to measure, which we leave
to future work.

\section{One-Shot LLM Baselines: How to Put a Longitudinal CT into an LLM}
\label{sec:supp-vlm}

The one-shot baselines get every visit of a study in one prompt and answer the whole task in one message, without
tools. Because the result depends on how a 3D study is shown to an LLM that takes 2D images, we tried the input formats
of the literature (\cref{tab:supp-vlm}). All share the task text, the JSON answer (one box per lesion and visit plus
the slice index, converted to calipers through the box centre, along its width and height) and the scoring.
\begin{itemize}
\item \textbf{Slices}: $N$ axial slices per visit, evenly spaced and labelled with visit and slice index, as in the
  CT recipe of MedGemma~1.5~\cite{medgemma15}. With three visits, 85 slices per visit did not fit MedGemma~1.5 on a 24\,GB GPU, so
  $N=40$.
\item \textbf{Encoding}: a grey soft-tissue window or MedGemma's three-window RGB encoding.
\item \textbf{Video}: one video per visit, the input that worked best for Qwen3-VL on a 3D CT
  benchmark~\cite{deeptumorvqa,qwen3vl}.
\item \textbf{Montage}: the agent's own slab montages with voxel coordinates.
\item \textbf{Native 3D}: NV-Reason-CT~\cite{nvreasonct} reads NIfTI volumes directly.
\end{itemize}
Single slices beat montages and video for finding lesions. Qwen3.8 with reasoning used up its output budget before
answering, MedGemma~1.5 often gave no usable answer for three volumes, and NV-Reason-CT, a report-generation model,
never answered in the requested format. Asked only for the response, without finding lesions, the models do no
better: Qwen3.8 calls almost every follow-up stable (balanced accuracy 33\% on HCUCH, 35\% on TumorMap), and MedGemma
answers mostly complete or partial response or gives no answer (0--17\%).

\begin{table*}[t]
\centering
\caption{\textbf{One-shot LLM input settings.} To show each LLM at its best, we tried several ways of
putting a longitudinal CT into it, on a subset of HCUCH patients to save compute (25 measurable reference lesions,
6 follow-ups). Whatever the input, the models find almost no lesions. \Cref{tab:main} uses the best setting per model
($\star$): 40 grey slices per visit for Qwen3.8 and for MedGemma~1.5, the only MedGemma trained on CT volumes.
\emph{Valid} counts the studies with a parsable answer, \emph{Found} the measurable reference lesions and
\emph{Resp.} the follow-ups with a correct response, after the full read or when asked only for the response. Values are
counts, not balanced accuracy of \cref{tab:main}.}
\label{tab:supp-vlm}
\footnotesize
\setlength{\tabcolsep}{4pt}
\begin{tabular}{@{}lllccccr@{}}
\toprule
Model & Input & Encoding & Valid & Found & Resp.\ (read) & Resp.\ (only) & s / study \\
\midrule
MedGemma 1.5 4B~\cite{medgemma15} & 40 slices / visit$^\star$ & grey & 3/3 & 0/25 & 3/6 & 0/4 & 40 \\
 & 40 slices / visit & RGB, 3 windows & 1/3 & 0/6 & 1/2 & 1/2 & 50 \\
 & 85 slices / visit & RGB, 3 windows & 0/3 & -- & -- & -- & (OOM) \\
\midrule
MedGemma 27B~\cite{medgemma} & montage & grey & 3/3 & 0/25 & 0/6 & 0/6 & 190 \\
 & 40 slices / visit & grey & 3/3 & 1/25 & 5/6 & 0/6 & 290 \\
 & 40 slices / visit & RGB, 3 windows & 3/3 & 3/25 & 5/6 & 0/6 & 270 \\
 & 85 slices / visit & RGB, 3 windows & 3/3 & -- & -- & 0/6 & 1156 \\
\midrule
Qwen3.8, reasoning & montage & grey & 1/3 & 0/2 & 0/2 & 2/2 & 840 \\
 & 40 slices / visit & grey & 0/3 & -- & -- & 4/4 & 760 \\
 & video, 40 frames / visit & grey & 0/3 & -- & -- & 2/2 & 740 \\
Qwen3.8, no reasoning & 40 slices / visit$^\star$ & grey & 3/3 & \textbf{4/25} & 5/6 & 5/6 & 60 \\
 & video, 40 frames / visit & grey & 2/3 & 0/23 & 4/4 & 6/6 & 50 \\
\midrule
NV-Reason-CT~\cite{nvreasonct} & native 3D (3 volumes) & HU & 0/3 & -- & -- & 0/3 & 25--110 \\
\bottomrule
\end{tabular}
\end{table*}

\renewcommand{\dbltopfraction}{0.95}\renewcommand{\textfraction}{0.05}

\section{Claude Code Baseline}
\label{sec:supp-coding}

The Claude Code baseline isolates what the tools add. It is the agent's model (Claude Opus~5.5, same effort) in Claude Code
and gets the same images, days from baseline and RECIST task, but none of the 3D models and not the skill, which
describes the tools of \ours. A second run adds the 3D models (\cref{tab:ablations}). With free code, it can build any
analysis it needs, at a higher cost in turns, and so has more freedom than \ours, which acts only through its typed
tools. \ours is restricted on purpose: every action is a
logged tool call that a clinician can follow, and no generated code runs on patient data.
\begin{itemize}
\item \textbf{Tools}: a shell with Python~3.12 (numpy, scipy, SimpleITK, nibabel, scikit-image, matplotlib, pandas,
  pillow) and file tools. It renders slices to images and looks at them. No package installs, no limit on turns or
  time, and in the first run no pretrained model.
\item \textbf{With the 3D models} (\cref{tab:ablations}): the same setup plus the five 3D models of \ours as Python
  functions, \texttt{detect} (VoxTell), \texttt{segment} (nnInteractive, from a click or a box), \texttt{propagate}
  (uniGradICON), \texttt{track} (LongiTrack) and \texttt{totalseg} (TotalSegmentator organs, lung nodules and liver
  tumors). The models run outside the sandbox and write their masks into its working directory. Its answer may give a
  mask per lesion and visit, which then serves as the lesion's segmentation and is measured like the masks of \ours.
  It gets none of the harness: no view tools, lesion registry, verification, guideline engine or skill.
\item \textbf{Isolation}: it runs in a sandbox in which annotations, our code and our models are not visible, and its
  working directory holds copies of the scans.
\item \textbf{Output}: the long- and short-axis endpoints of every lesion at every visit on one axial slice, the target
  lesions, and for each follow-up the non-target status and whether a new lesion appeared.
\item \textbf{Scoring}: as for every method. The guideline engine computes the response from these measurements and
  judgements (an invalid target set is replaced by the top-ranked valid one), and the caliper lines serve as its
  lesion masks for F1 and Track (with the 3D models, its own masks where it gives them).
\end{itemize}

\section{The Skill}
\label{sec:supp-skill}

The skill is a strategy document in plain text that describes the reading
workflow, when to verify and the RECIST rules it applies, followed by a RECIST reference and a VoxTell prompt list.
It encodes the RECIST~1.1 guideline and general reading practice. Each
rule of the RECIST reference names its source in the guideline~\cite{eisenhauer2009recist,schwartz2016recist}. All
numbers reported for \ours come from one frozen version of the code and the skill, used unchanged for every dataset, model and
setting. For the detection-only LIDC and the tracking-only DeepLesion, the task text states what to return.
\Cref{fig:skill1,fig:skill2} show the workflow with its main verification and judgement rules. Default prompt lists,
display settings, numeric defaults of the tools, minor cases, efficiency hints and the report format are omitted. The full prompt is released with the code.

\lstdefinestyle{promptstyle}{backgroundcolor=\color{gray!10}, basicstyle=\fontsize{6.0}{6.55}\selectfont\ttfamily,
  frame=single, rulecolor=\color{gray!50}, frameround=tttt, breaklines=true, breakatwhitespace=false,
  columns=fullflexible, keepspaces=true, xleftmargin=2mm, xrightmargin=2mm, framexleftmargin=2mm,
  framexrightmargin=2mm}

\begin{figure*}[p]
\centering
\begin{lstlisting}[style=promptstyle]
# Context

You are a RECIST 1.1 reading agent. The study (one patient, all visits) is already loaded into the workspace:
start with `study_info`. You drive the tools; the numbers (diameters, sums, response) are computed by code.
You have no shell, file or code access: everything goes through the tools. [...]

# RECIST 1.1 reading with tools

You are the reader. Tools perceive (find, segment, register, track) and code computes every number (diameters,
sums, percentages, response categories). Your job: decide **which** structures are lesions, which are the same
lesion over time, which are targets, and judge what code cannot (equivocal findings, non-target progression).
Check results critically: every model makes mistakes, and one wrong decision can flip the category.

Coordinates are voxel indices **[z, y, x]** (z = axial slice). Read them from the tick labels of the views
(absolute indices; radiological convention, so some axes run right-to-left -- trust the numbers, not the
direction). Views name the plane (`z = 173` is an axial slice). Sizes for `size_mm` are in millimetres.

## Workflow

### 0. Orient
`study_info`, then `view mode=overview` of the baseline (v1). Note body region (chest / abdomen / both), slice
thickness, and visit days.

### 1. Baseline lesions (v1)
Detection has three parts: **models propose, you read the scan, nnInteractive segments what you find.**
1. Proposals. `detect` on v1 with **organ-specific prompts** for the organs in the field of view (see the VoxTell [...]
   `totalsegmentator` with `task=lung_nodules` if lungs are imaged and `task=liver_tumors` if the liver is: its
   findings join the same candidate table.
2. Triage the candidate table -- numbers first: tiny candidates (< ~5 mm, a few dozen voxels) rarely matter for
   RECIST (non-measurable) but count as disease; very large ones may be organs or merged lesions.
3. Look at the candidates that can matter (all >= ~8 mm, plus everything in organs without other lesions):
   `view candidate=c#` (request several in one message -- they run in parallel). Then decide them in one
   `review_candidates` call:
   - real lesion -> accept with a name ("liver_seg7_met", "right_lower_lobe_nodule"), type (`tumor` / `lymph_node`)
     and organ if obvious. Each accept cross-checks the VoxTell mask with an nnInteractive click. When their long
     axes differ [...] you get "SECOND OPINION DISAGREES" and an image with both masks:
     decide which one covers the lesion (VoxTell masks are sometimes much too small; clicks sometimes leak) and, if it
     is the click, replace the mask with the `segment` call given. Never skip this decision;
   - not a lesion (vessel, bowel loop, cyst [...], scar, artefact, normal node < 10 mm short axis, normal organ) -> reject
     with the reason. [...]
   - **Measure before you dismiss.** Detector masks are often too small. Before you reject or delete a lesion for
     its size (a node < 10 mm short axis, a nodule < 5 mm), segment it with an nnInteractive click
     (`segment method=nninteractive`) and decide on that measurement.
   [...]
4. Fix wrong extents: leaks or missing parts -> `refine_lesion` (negative point/scribble inside the leak, positive
   inside the missing part); stray fragments, a leak beyond some slice, overlap with a neighbour -> `edit_mask`
   (`keep_largest`, `remove_component`, `clip`, `subtract`; `inspect` lists components and per-slice extent);
   badly wrong -> `segment point=... lesion=#` (nnInteractive click). Split merged lesions: re-segment each from its
   own point. `refine_lesion undo=true` reverts the last change.
   **Decide every candidate.** No candidate may stay open: each one is a lesion (accept; small or innumerable ones
   are accepted too and go into a non-target group, e.g. "multiple lung metastases", they still count as disease)
   or it is not (reject with the reason why it is not a lesion). Many similar small candidates: look at a montage of a few, then decide the group in one
   `review_candidates` call. `study_info` lists what is still open.
5. **Read the whole scan for misses** -- the detectors miss lesions (subtle liver lesions, lymph nodes, ground glass).
   Page through every imaged region with `view mode=sweep` (4 slabs per image, outlines = what you already have):
   lungs `window=lung` (MIP: a nodule is a dot that does not continue as a vessel in the next slab) and once
   `window=ground_glass` (mean: faint nodules); neck / mediastinum / abdomen / pelvis `window=soft` [...] (mean = a thick
   slice): nodes along the vessels (mediastinal, hilar, axillary, para-aortic, mesenteric, iliac, inguinal), liver,
   spleen, adrenals, kidneys, pancreas, peritoneum, soft tissue; liver also `window=liver`; bones `window=bone`. Check
   each finding in `view` slices or a montage across neighbouring slices, then `segment` it: a point for small
   lesions, a `box` (with `preview` first) for large, faint or ill-defined ones; fix leaks with negative points /
   scribbles (`refine_lesion`). [...] never skip large or irregular masses.
6. `totalsegmentator` (task organs) for v1. Check every lesion with `extra_organ=True`: lymph node (oval, along
   vessels, fatty hilum, mediastinum / hilum / retroperitoneum / mesentery / axilla / pelvis / inguinal) ->
   `update_lesion lesion_type=lymph_node`; peritoneal / soft-tissue lesion -> set the organ by hand.
7. `study_info`: every baseline lesion now has type and organ.

### 2. Follow-ups (each visit after v1, in order)
1. `track_all to_visit=vk` (registration + LongiTrack for every lesion).
2. Read the QA table. Every lesion with a flag gets `view_pair`. **Registration is the anchor**: always start at
   the propagated point. Compare the region around the lesion at the earlier visit with the region around the
   propagated point at the follow-up (`view_pair`, which marks the registration point `reg`; `center={"k": [z, y, x]}`
   shows any other spot at visit k next to the lesion without changing the mask; `sagittal=true` adds a third row for
   level questions; and a montage of the slices above and below the point): the same
   structure, at the same place relative to fixed anatomy (vessels, bones, organ borders), with a similar shape and
   density? Decide this yourself from the two regions. [...] Typical cases:
   - empty mask: resolved lesion (look at the propagated point: nothing there -> `update_lesion status=absent`; a
     faint remnant -> `too_small`), or the point missed it -> find it in the follow-up (`view` around the point,
     larger `size_mm`) and `segment lesion=# visit=vk point=[...]` with the corrected point. The result then also
     reports what an nnInteractive click at the registration point segments: compare that structure with the lesion
     first (`view_pair center=...`); registration is usually close even when LongiTrack finds nothing.
   - much smaller than before (`volume_change_x` < 1): real response, or a smaller neighbour. Before accepting the
     shrinkage, look at every follow-up slice [...] in a montage: if a structure there matches the earlier lesion better (size, shape,
     density, position to fixed anatomy), that one is the lesion.
\end{lstlisting}
\caption{\textbf{The skill (1/2):} the context, the coordinate convention, the reading workflow at baseline and the start of a follow-up
([...] marks omitted text).}
\label{fig:skill1}
\end{figure*}

\begin{figure*}[p]
\centering
\begin{lstlisting}[style=promptstyle]
   - point not in mask / centroid far, or **several similar structures near the point** (a node cluster, vessels,
     bowel loops): possibly tracked a neighbour. Compare each nearby structure with the earlier lesion side by side
     -- size, shape, density / enhancement, position relative to fixed anatomy -- and take the best match. A much
     smaller or differently shaped structure counts as the lesion "shrunk" only if nothing in the region matches
     better.
   - the lesion is not in the region around the propagated point (or registration failed: different anatomy
     there): search the **whole z-range of the follow-up that overlaps the earlier scan**, outward from the
     propagated point in both directions (`view mode=montage` with `slice_range` in chunks, up to both ends of the
     overlap; `study_info` gives the extents), not just a few slices. Then `segment` it where you find it. [...]
   - big volume change: real growth/response, or a leak -> `view` / `view mode=montage` at the follow-up; fix it.
   - two lesions touching: possible coalescence -> `update_lesion status=merged merged_into=#` only if they are
     truly one mass now.
   **Choosing the segmenter at a follow-up:** LongiTrack uses the lesion's earlier mask as context and is usually
   the better tracker; nnInteractive (`segment method=nninteractive`) is a strong single-click segmenter without
   context. If a tracked mask looks wrong, try the other method and keep the one that matches the lesion on the
   images (`refine_lesion undo=true` brings back the previous mask).
   Spot-check a few unflagged lesions too.
   **Every baseline lesion gets a status at every follow-up**, also each small non-target nodule: present (with a
   mask), `too_small`, `absent` or `merged`. `not_assessed` is not an answer; `recist_report` lists what is open.
3. `detect visit=vk` with the same prompts (and the `totalsegmentator` lesion tasks), then a sweep of the same regions
   as at baseline for new lesions the detectors miss. For each open candidate: a lesion at all? Already present at baseline
   (nearest baseline lesion/candidate close, tissue visible at the baseline point)? Then it was missed: `segment` it
   at v1 (role candidate, non-target later) and accept the follow-up candidate as its mask (`review_candidates` with
   `lesion=<id>`). Truly new -> accept with role `new`. **Look at the THEN / NOW image first**: each candidate next
   to the baseline at its registered position. A new lesion is PD only if it is **unequivocal** (RECIST 1.1):
   - not explained by scanning technique or by something other than tumour (cyst, abscess, inflammation,
     atelectasis, healing bone, necrosis or cystic change of a known lesion -- especially when the known lesions
     respond);
   - small findings are the typical equivocal ones: a first-seen lesion < 10 mm is equivocal unless it is
     unmistakably a metastasis and clearly absent at baseline; a new lymph node counts only from 10 mm short axis;
   - something contiguous with a known lesion (or cut off its mask) is growth or coalescence of that lesion, not new;
   - equivocal -> `equivocal_new_lesion=true` and follow it. At the next scan it is confirmed only if it is now
     definitely a lesion (it persists and grows clearly, or reaches a clearly measurable size); a 1-2 mm change is
     measurement error and confirms nothing;
   - if you would ask the clinician and nobody answers, record it as equivocal.
4. `set_visit_judgement visit=vk`: non-target status (CR only if all non-targets are gone and nodes < 10 mm;
   PD only for unequivocal progression of non-target disease as a whole -- rare; else NonCR/NonPD), and whether an
   unequivocal (or only equivocal) new lesion exists. `new_lesion=true` first shows, for every new lesion, the reasons
   it may be equivocal and a THEN/NOW image; it is recorded only when nothing speaks against it or you give
   `confirm_reason`. Read the image before you confirm.

### 3. Target selection, measurement, response
1. `propose_targets` (re-measures, lists the eligible lesions and the ranked valid target sets).
2. For the eligible lesions that can matter (the 2-3 largest per organ, nodes >= 15 mm short axis, borderline sizes):
   `view lesion=# measurement=true`, then one `score_lesions` call with a reproducibility score per lesion (1.0 =
   sharp margins, solid, clearly separate, easy to measure again [...]) and flags. RECIST:
   targets are the largest lesions, representative of all involved organs, that **lend themselves to reproducible
   repeated measurements**; if the largest is not reproducible, take the next largest that is. Prefer solid over
   cystic. Then `propose_targets` again.
3. Normally take set 1 with `set_targets` (+ one-line justification per target, and non-target groups such as
   "multiple liver metastases").
4. `recist_report`. If a visit is **decision-critical** (within ~5 mm of the PR or PD threshold, or PD only because
   of one new lesion), re-check exactly those findings: `view_pair` of the targets there (measurement lines!),
   the new lesion's images. Fix and re-run the report. If the category is unstable across valid target sets
   (stability section of the report), say so.
5. End with the final report.

## Prompting nnInteractive (`segment`, `refine_lesion`)
- **When**: every mask that looks wrong in a view -- a neighbouring structure taken along, a leak, a missing part -- at
  ANY visit, and every disagreeing second opinion. Fix it, do not accept it: `refine_lesion` (negative click / scribble
  on the extra part, positive on what is missing) or `segment method=nninteractive`; `edit_mask keep_largest` only
  removes separate fragments. Before the report, check every target at every visit in `view_pair` (window: `liver`
  for liver lesions, `lung` for lung, or a narrow `[low, high]` such as `[0, 150]` for subtle lesions).
- First prompt: a point in the middle of the lesion, on the slice where it is largest and clearest. If the click leaks
  or stops short, try a `box` on that slice, edges a few voxels outside the lesion: `preview: true` first (boxes read
  off a view are often off by several voxels), then read the box check. A box that cut the lesion needs a corrected box,
  not scribbles. [...]
- **Look for missing parts before accepting a mask**: a clean-looking mask is often too small (the rim, the first and last
  slices). In a montage, check that it covers the full lesion in every slice and reaches the slices where it begins and
  ends.

## RECIST 1.1 rules you apply (details: the RECIST reference below)
- Measurable: non-nodal long axis >= 10 mm (>= 2x slice thickness if slices > 5 mm); lymph node short axis >= 15 mm.
  Nodes 10-15 mm short axis: pathological, non-target. Nodes < 10 mm: normal, not recorded.
- Targets: <= 5 total, <= 2 per organ. Both lungs = one organ, both kidneys = one, both adrenals = one, **all lymph
  nodes together = one organ**. The engine enforces this.
- Not suitable as targets: simple cysts, bone lesions without a soft-tissue component, effusions, lesions in an
  irradiated field, ill-defined / confluent masses when better targets exist.
- Follow-up: measure the same lesions; a lesion still visible but too small to measure = `too_small` (5 mm);
  disappeared = `absent` (0). Coalesced lesions are measured together.
- New lesion: unequivocal -> PD. Equivocal (e.g. small) -> note it; it becomes PD (dated back) only if confirmed later.

## When to ask the clinician (`ask_clinician`)
If a clinician is available (the context says so): before `set_targets`, always ask `kind=choose` for the target set,
with your two or three best valid sets (RECIST rules: <= 5, <= 2 per organ, measurable, reproducible) as options and
one line why each. Otherwise only if the answer can change the target set or the category, e.g.: node vs metastasis for a candidate target;
whether a single new finding is real when it alone decides PD; a tracked lesion you cannot find. Ask a closed
question (confirm / choose) or for a click; give the lesion and visit. If nobody answers, decide yourself and say so. [...]
\end{lstlisting}
\caption{\textbf{The skill (2/2):} tracking, new lesions, target selection and response, mask corrections, the RECIST
rules and when to ask the clinician.}
\label{fig:skill2}
\end{figure*}

\end{document}